\documentclass[preprint,12pt,review]{elsarticle}

\usepackage{fullpage} 
\usepackage{pdflscape}
\usepackage{afterpage}

\usepackage[english]{babel}
\usepackage{lineno}
\usepackage{amsmath}
\usepackage{amssymb}
\usepackage{multirow}
\usepackage{color}
\usepackage[usenames,dvipsnames]{xcolor}
\usepackage{algorithm}
\usepackage[noend]{algpseudocode}
\usepackage{arydshln} 
\usepackage{caption}
\usepackage{subcaption}

\usepackage{placeins}

\usepackage{hyperref}

\def\Unet{U\nobreakdash-net\, }

\journal{Computer Methods and Programs in Biomedicine}

\biboptions{numbers,sort&compress}

\begin{document}

\begin{frontmatter}



\title{Automatic quantification of the LV function and mass: a deep learning approach for cardiovascular MRI}


\author[Affil1]{Ariel H. Curiale \corref{cor1}}
\ead{ariel.curiale@cab.cnea.gov.ar (http://www.curiale.com.ar)}
\author[Affil2,Affil3]{Flavio D. Colavecchia}
\author[Affil1,Affil3]{German Mato}

\address[Affil1]{CONICET -  Departamento de F\'isica M\'edica, Centro At\'omico Bariloche, Avenida Bustillo 9500, 8400 S. C. de Bariloche, R\'io Negro, Argentina}
\address[Affil2]{CONICET - Centro Integral de Medicina Nuclear y Radioterapia, Centro At\'omico Bariloche, Avenida Bustillo 9500, 8400 S. C. de Bariloche, R\'io Negro, Argentina.}
\address[Affil3]{Comisi\'on Nacional de Energ\'ia At\'omica (CNEA).}

\cortext[cor1]{Address correspondence to: Ariel Hern\'an Curiale, 
Departamento de F\'isica M\'edica, Centro At\'omico Bariloche, Avenida Bustillo 9500, 8400 S. C. de Bariloche, R\'io Negro, Argentina.}

\begin{abstract}
Cardiac function is of paramount importance for both prognosis and treatment of different pathologies such as mitral regurgitation, ischemia, dyssynchrony and myocarditis. Cardiac behavior is determined by structural and functional features. In both cases, the analysis of medical imaging studies requires to detect and segment the myocardium. Nowadays, magnetic resonance imaging (MRI) is one of the most relevant and accurate non-invasive diagnostic tools for cardiac structure and function. 
In this work we use a deep learning technique to assist the automatization of left ventricle function and mass quantification in cardiac MRI. We study three new deep learning architectures  specially designed for this task  where the generalized Jaccard distance is used as optimization objective function. Also, we integrate the idea of sparsity and depthwise separable convolution into the \Unet architecture, as well as, a residual learning  strategy level to level. Our results demonstrate a suitable accuracy for myocardial segmentation ($\sim0.9$ Dice's coefficient), and a strong correlation with  the most relevant functional measures: 0.99 for end-diastolic and end-systolic volume, 0.97 for the left myocardial mass, 0.95 for the ejection fraction and 0.93 for the stroke volume and cardiac output. It is important to note that the errors are comparable to the inter e intra-operator ranges for manual contouring. 

\end{abstract}


\end{frontmatter}



\section{Introduction}
\label{intro}

Some of the most relevant global structural features for quantification of the cardiac function are the left ventricular mass (LVM), the left ventricular volume (LVV) and the ejection fraction (EF), which is directly derived from the left ventricle (LV) at end-diastole (ED) and end-systole (ES).
Left ventricle function and mass are of paramount importance for both prognosis and treatment of different cardiac pathologies such as mitral regurgitation, ischemia and myocarditis~\citep{Lowes1999,Koelling2002,Friedrich2009,Edvardsen2002,Suffoletto2006}. For instance, LVM is considered as an independent predictor of cardiovascular events, while LVV is associated with adverse remodeling~\citep{Bluemke2008s,Gjesdal2011w,Suinesiaputra2015s}.
%
%
Cardiovascular magnetic resonance  (CMR) is one of the most accurate non-invasive diagnostic tools for imaging of cardiac structure and function~\citep{Weinsaft2007}. Usually, it is considered as the gold standard for LV  mass and volume quantification~\citep{Gerche2013s}.  In this way, manual or semi-automatic delineation by experts is currently the standard clinical practice for  chamber segmentation from CMR images. This step is essential  for quantification of global features, such as the estimation of ventricular volume, ejection fraction and myocardial mass. Despite the efforts of researchers and medical vendors, global quantification and volumetric analysis still remain time consuming tasks which heavily rely on user interaction. For example, a diastolic functional evaluation performed by using dedicated software with manual segmentation of basal section commonly takes 25 minutes by an experienced radiologist~\citep{Graca2014k}.  Thus, there is still a significant need for tools that allow  automatic 3D quantification. The main goal of this work is to provide an accuracy and  suitable  approach for LV function and mass quantification for CMR based on deep convolutional neural networks.
%

%
Recently, convolutional neural networks (CNNs)~\citep{Lecun1998s} have been successfully used for solving challenging tasks like  classification, segmentation and object detection, achieving state-of-the-art performance~\citep{Litjens2017k}.
%
In fact, fully convolutional networks trained end-to-end have been recently used for medical images, for example, for cell, prostate and myocardial tissue segmentation~\citep{Ronneberger2015, Milletari2016p, Curiale2017l}. These models, which serve as an inspiration for our work, employ different types of network architectures and were trained to generate a segmentation mask that delineates the structures of interest in the image. Which in our case are the myocardial tissue and blood pool for the LV.

Deep neural networks, such as CNN, are very useful tools for pattern recognition, however,  one  of the main disadvantages arises from the complexity of their training stage. For instance, the distribution of each layer's inputs changes along this process, as the parameters of the previous layers change. This phenomenon is called  internal covariate shift~\citep{Shimodaira2000}.
As the networks start to converge, a degradation problem occurs: with increasing network depth, accuracy gets saturated. Unexpectedly, such degradation is not caused by overfitting, and adding more layers to a suitably deep model leads to higher training error, as it was reported by~\cite{He_2015_CVPR}.
To overcome this problem the technique of batch normalization has been proposed~\citep{Ioffe2015k}. This procedure involves the evaluation of the statistical properties of the neural activations that are present for a given batch of data in order to normalize the inputs to any layer to obtain some desired objective (such as zero mean and unit variance). At the same time, the architecture is modified in order to prevent loss of information that could arise from the normalization. This technique allows to use much higher learning rates and also acts as regularizer,  in some cases eliminating the need for dropout~\citep{Srivastava2014}.

Another recent performance improvement of deep neural networks has been achieved by reformulating the layers as learning residual functions with reference to the layer inputs, instead of learning unreferenced functions~\citep{He2016z}. In other words, the parameters to be determined at a given stage generate only the difference (or residual) between the objective function to be learned and some fixed function such as the identity. It was empirically found that this approach gives rise to networks that are easier to optimize, which can also gain accuracy from considerably increased depth~\citep{He2016z}. 

The most straightforward way of improving the accuracy of deep neural networks is by increasing the number of levels (deep size) and units by level (width size). In this way, it is possible to train  higher quality models. However, bigger size implies an increase of computational resources because  a large number of parameters have to be trained which ends, among other things, in overfitting.
Both issues can be overcome by introducing the notion of sparsity as it was proposed in the Inception network~\citep{Szegedy2015v}.
The fundamental hypothesis behind Inception is that cross-channel correlations and spatial correlations are sufficiently decoupled that it is preferable not to map them jointly as it was pointed out in the Xception architecture~\citep{Chollet2017z}. Indeed, the Xception is based entirely on depthwise separable convolution layers which allows to extremely reduce the network complexity regarding to the number of trainable parameters, but maintaining or improving the generalization power.

In this work, we explore the idea of information sparsity described in the Inception, and also, the notion of depthwise separable convolutions introduced by \cite{Chollet2017z}. Both ideas are introduced into a \Unet architecture~\citep{Ronneberger2015} for myocardial tissue classification and three new deep learning networks are analyzed for this task. Then, the most accurate approach for myocardial tissue classification is used for quantification of the LV function and mass.
Unlike previous works for myocardial tissue classification~\citep{Curiale2017l}, we propose to use the  generalized Jaccard distance~\citep{Crum2006r} as optimization objective to properly handle fuzzy  sets.
Results demonstrate that the our approach outperforms previous methods based on the \Unet architecture~\citep{Curiale2017l} and provides a suitable automatic approach for myocardial segmentation and cardiac function quantification. 

The paper is structured as follows: in Section 2 the fully automatic approach based on the \Unet network is introduced. Additionally, the training strategy and the optimization objective function is presented. In Section 3,  the idea of information sparsity, depthwise separable convolutions and residual learning level to level are evaluated for myocardial tissue classification. Finally, we present the conclusions in Section 4.
%

\section{Materials and methods}

\subsection*{Materials}

The proposed approach for automatic quantification of the LV function and mass was trained and evaluated with 140 patients from two public CMR datasets: the Sunnybrook Cardiac Dataset (SCD)~\citep{Radau2009h} and the  Cardiac Atlas Project (CAP)~\citep{Fonseca2011r}.

The SCD dataset, also known as the 2009 Cardiac MR Left Ventricle Segmentation Challenge data, consists of 45 cine-MRI images from a mix of patients and pathologies: healthy, hypertrophy, heart failure with infarction and heart failure without infarction. A subset of this dataset was first used in the automated myocardium segmentation challenge from short-axis MRI, held by a MICCAI workshop in 2009. The whole complete dataset is now available in the Cardiac Atlas Project dataset with public domain license\footnote{\url{http://www.cardiacatlas.org/studies/sunnybrook-cardiac-data}}.
The 45 cardiac cine-MR were acquired as cine steady state free precession (SSFP) MR short axis (SAX) with 1.5T General Electric Signa MRI. All the images were obtained during 10-15 second breath-holds with a temporal resolution of 20 cardiac phases over the heart cycle, a mean spatial resolution of 1.36~ x~136~x~9.04~mm (255~x~255~x~11~ pixels), and scanned from the ED phase.  In these SAX MR acquisitions, endocardial and epicardial contours were drawn by an experienced cardiologist in all slices at ED, and only endocardial contours were provided at ES. All the contours were confirmed by another cardiologist.

The Cardiac Atlas Project provides a set of CMR for 95 patients from a prospective, multi-center, randomized clinical trials in patients with coronary artery diseases and mild-to-moderate left ventricular dysfunction.
All cines SSFP were acquired during a breath-hold of 8-15 seconds duration with a typical thickness $\le$ 10 mm, gap $\le$ 2 mm, TR 30-50 ms, TE 1.6 ms, flip angle $60^{\circ}$, FOV 360 mm, and mean spatial resolution of 1.48~x~1.48~x~9.3~mm (245~x~257~x~12~ pixels). Sufficient short-axis slices were acquired to cover the whole heart in SAX. Also, in these acquisitions the myocardial manual segmentation were provided.

\subsection*{Methods}
\label{sec_method}
The method described in this section was intentionally designed to measure the LV function and mass by means of identifying  the myocardial tissue an the blood pool for the LV as it is shown in Fig.~\ref{fig_workflow}.  To this end, a deep learning approach is introduced for detecting a proper region of interest (ROI) around the LV, and also, for myocardial tissue and blood pool classification. Then, the LV function and mass is derived as follows~\citep{Frangi2001l}:

\begin{figure*}[!t]
\centering
\includegraphics[width=\textwidth]{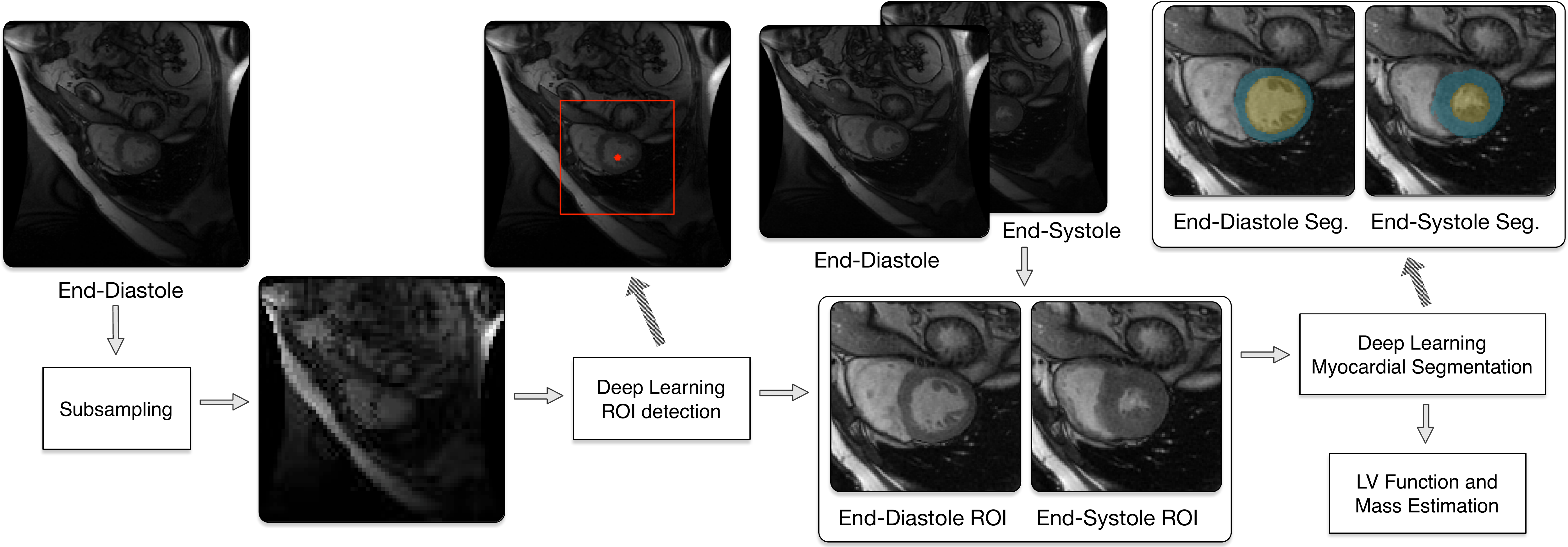}
\caption{Workflow of the proposed method. \label{fig_workflow}}
\end{figure*}

\begin{itemize}
\item
\textbf{Left ventricle mass (LVM):}  is directly derived from the myocardial segmentation by making two main assumptions: (a)  the interventricular septum is assumed to be part of the LV and (b) the myocardial volumen is equal to the total volume contained within the epicardial borders of the ventricle, $V_{\text{epi}}$, minus the chamber volume, $V_{\text{endo}}$ at end-diastolic frame ($t_{\text{ED}}$):           
$$\text{LVM} =  \rho \cdot ( V_{\text{epi}}(t_{\text{ED}}) -V_{\text{endo}}(t_{\text{ED}})) \, ,$$
where $\rho = 1.05 \ \text{g}/\text{cm}^3$ corresponds to the myocardial tissue density. LVM is usually normalized to total body surface area or weight in order to facilitate interpatient comparisons. 

\item 
\textbf{Stroke volume (SV):} is defined as the volume ejected between the end of diastole and the end of systole:
$$\text{SV} =   V_{\text{endo}}(t_{\text{ED}}) - V_{\text{endo}}(t_{\text{ES}}) \, ,$$

\item 
\textbf{Cardiac Output (CO):} The blood flow which is delivered by the heart with oxygenated blood to the body is known as the cardiac output and is expressed in liters per minute.
$$\text{CO} =   \text{SV} \cdot h_r \, ,$$
where $h_r$ is the heart rate. Since the magnitude of CO is proportional to body surface, if an interpatient comparison is required, it should be adjusted by the body surface area.

\item 
\textbf{Ejection Fraction (EF):} is a global index which is generally considered as one of the most meaningful measures of the LV pump function. It is defined as the ratio between the SV and the end-diastolic volumen:
$$\text{EF} = \frac{\text{SV}}{ V_{\text{endo}}(t_{\text{ED}})} \times 100\%  \, .$$

\end{itemize}

\begin{figure*}[!t]
\centering
\includegraphics[width=\textwidth]{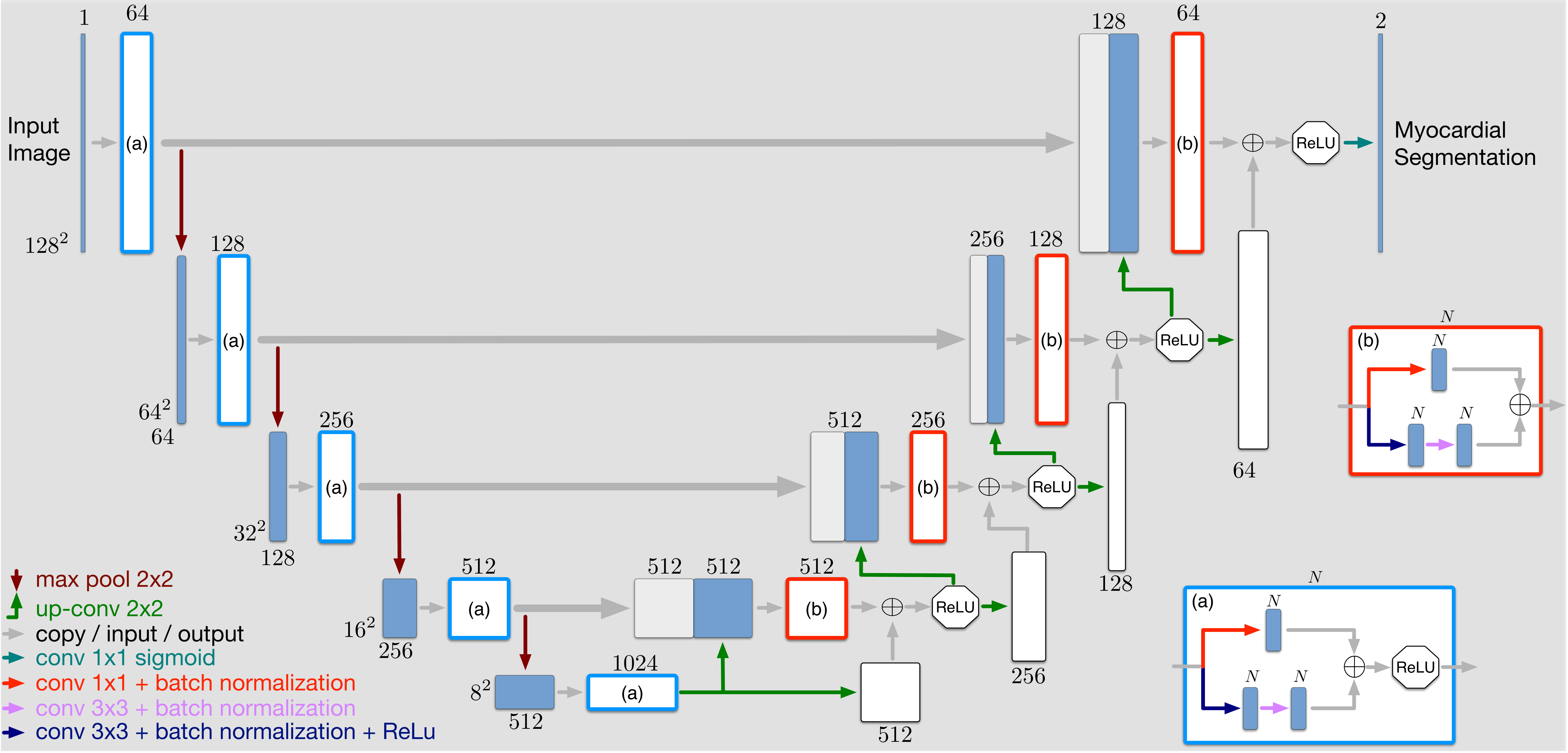}
\caption{Network Architecture proposed for myocardial segmentation in cardiac MRI. The number of channels/features is denoted on top of the box and the input layer dimension is provided at the lower left edge of the box. The arrows denote the different operations according to the legend. Withe blocks correspond to the differences with respect to previous approaches.\label{fig_architecture}}
\end{figure*}

The LV-ROI detection and blood/myocardial tissue classification are performed by using the same CNN approach  based on the \Unet. This CNN architecture is depicted in Fig.~\ref{fig_architecture}. The LV-ROI detection is carried out by using low spatial resolution images at end-diastole, while the blood pool and myocardial tissue classification is done by using the original information in the previously detected ROI around the LV (128 x 128 pixels with 190~x~190 mm). In particular, the ROI size was defined to ensure that the whole myocardial tissue will be covered.

In the proposed deep learning architecture, the network learns how to encode information about features presented in the training set (left branch on Fig.~\ref{fig_architecture}). Then, in the decode path the network learns about the image reconstruction process from the encoded features learned.

The specific feature of the \Unet architecture lies on the concatenation between the output of the encode path, for each level, and the input of the decoding path (denoted as big gray arrows on Fig.~\ref{fig_architecture}). These concatenations provide the ability to localize high spatial resolution features to the \emph{fully convolutional network}, thus, generating a more precise output based on this information. As mentioned in \cite{Ronneberger2015} this strategy allows the seamless segmentation of  large images by an overlap-tile strategy.

The encoding path consists of two 3~x~3 convolution, each followed by a batch normalization and a residual learning just before performing the 2~x~2 max pooling operation with stride 2 as it was described in~\cite{Curiale2017l}. In fact, the batch normalization is performed right after each convolution and before activation, following~\cite{Ioffe2015k}. Also, non rectified linear unit (ReLU) is applied right after the addition used for residual learning as it is depicted in Fig.~\ref{fig_architecture}. At each downsampling step we double the number of feature channels, that is initially set to 64.

Every step in the decoding path can be seen as the mirrored step of the encode path, i.e.  each step in the decoding path consists of an upsampling of the feature map followed by a 2~x~2 convolution (``up-convolution'') that halves the number of feature channels, a concatenation with the corresponding  feature map from the encoding path, and two 3~x~3 convolutions, each followed by a batch normalization and a residual learning. Finally, a residual output learning is introduced from the upsampled previous level (left side Fig.~\ref{fig_architecture}). At the final layer, a 1~x~1 convolution is carried out to map the 64 feature maps to the two classes used for the myocardial segmentation (myocardium and endocardium). 
The output of the last layer, after soft-max non-linear function, represents the likelihood of a pixel belongs to the myocardium or  endocardium of the left ventricle. Indeed, only those voxels with higher likelihood ($>0.5$) are considered as part of the left ventricle tissue.

\subsection*{Training}

The  CNN used for detecting a ROI around the LV was trained with low spatial resolution images and their corresponding manual segmentation (64~x~64 pixels) with a mean spatial resolution of 5.76~x~5.76 mm. In contrast, the blood pool and myocardial tissue classification was performed by training a CNN with cropped images in short-axis (128~x~128 pixels) according to the previously LV-ROI detection with a mean spatial resolution of 1.44~x~1.44 mm.
A 5 fold cross-validation strategy is used for training the CNN's due to the reduced number of patients. In this way, 5 sets of training/validation were used for each dataset. A total of 4048 2D images were used for training  (36 patients from SCD and 76 patients from CAP)  and 1003 2D images for testing (8 patients from SCD and 19 patients from CAP). 
%
In both CNN's the optimization was carried out by using the stochastic gradient descent (Adaptive Moment Estimation) implementation of Keras~\citep{Chollet2015d} with a learning rate of $10^{-4}$.
Also, the generalized Jaccard distance~\citep{Crum2006r} is used as the loss objective function to properly handle fuzzy  sets:
\begin{eqnarray}
\text{J}_d(X, Y) =&  1 - \frac{\sum_i  \min(X_i, Y_i)}{\sum_i  \max(X_i, Y_i)} ,  \nonumber \\
\end{eqnarray}
where $X$ and $Y$ are the myocardial segmentation prediction and the ground truth segmentation, respectively.

\begin{figure*}[!t]
\centering
\begin{subfigure}[b]{0.49\textwidth}
\includegraphics[width=.49\textwidth]{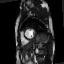}
\includegraphics[width=.49\textwidth]{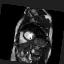}
\caption{\label{fig_da_roi}}
\end{subfigure}
\begin{subfigure}[b]{0.49\textwidth}
\includegraphics[width=.49\textwidth]{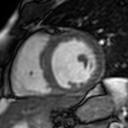}
\includegraphics[width=.49\textwidth]{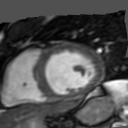}
\caption{\label{fig_da_classification}}
\end{subfigure}
\caption{Examples of the data augmentation used for training the CNN for one of the five cross-validation set for the ROI detection (a) and tissue classification (b).\label{fig_dataaugmentation}}
\end{figure*}

Annotated medical information like myocardial classification is not easy to obtain due to the fact that one or more experts are required to manually trace a reliable ground truth of the myocardial  classification. So, in this work it was necessary to augment the original training dataset in order to increase the examples from 4048 to 20000 2D images. 
Also, data augmentation is essential to teach the network the desired invariance and robustness properties. Heterogeneity in the cardiac MRI dataset is needed to teach the network some shift and rotation invariance, as well as robustness to deformations.
With this intention, the input of the network was randomly deformed by means of a spatial shift in a range of 10\% of the image size, a rotation in a range of $10^{\circ}$ in the short axis, a zoom in a range of 2x or by using a gaussian deformation field  ($\mu=0$ and $\sigma \in [0.7, 1.2]$ randomly chosen) and B-spline interpolation (Fig.~\ref{fig_dataaugmentation}).

\section{Results}

Three sets of experiments are conducted to evaluate the proposed methodology on both datasets (Sunnybrooks and CAP). First, the our approach was evaluated to measure the accuracy of the LV-ROI detection. Second, the accuracy on myocardial tissue classification was study for different CNN's. Finally, the third set of experiments were designed to measure the accuracy of the proposed automatic approach for quantification of the LV function and mass as it was described in Section~\ref{sec_method}.

In the experiments, the  papillary muscles (PM) were excluded because only the Sunnybrooks dataset contains this information.  So, the proposed method will avoid to detect the PM as part of the myocardial tissue.  The mean squared error and the Dice's coefficient were used to measure the accuracy of the proposed CNN for an easy comparison with others methods described in the bibliography.

\subsection*{ROI detection}

\begin{figure*}[!t]
    \centering
    \begin{subfigure}[b]{0.48\textwidth}
        \includegraphics[width=\textwidth]{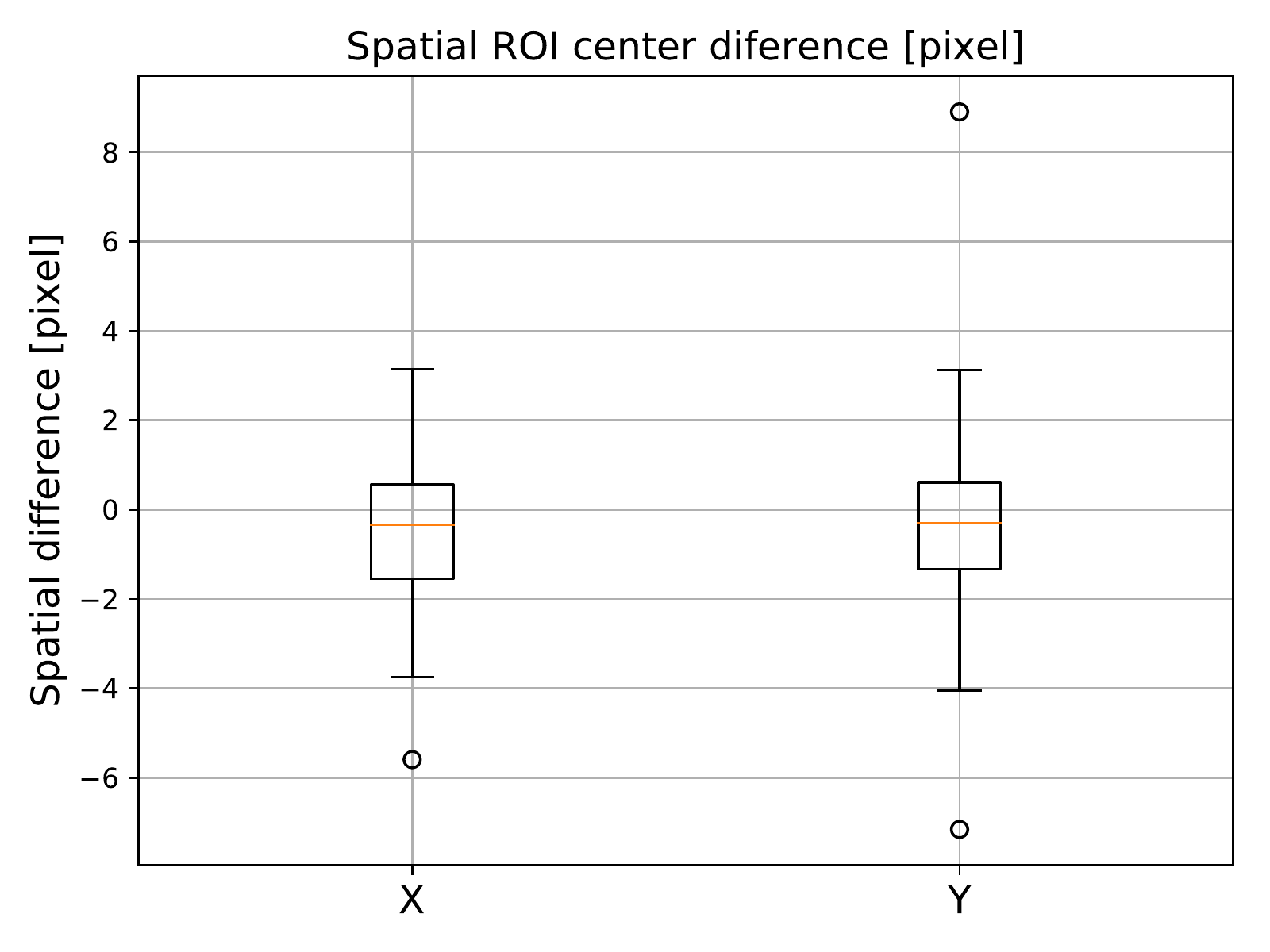}
        \caption{\label{fig_roi_diff}}
    \end{subfigure}
    \begin{subfigure}[b]{0.48\textwidth}
	\includegraphics[width=\textwidth]{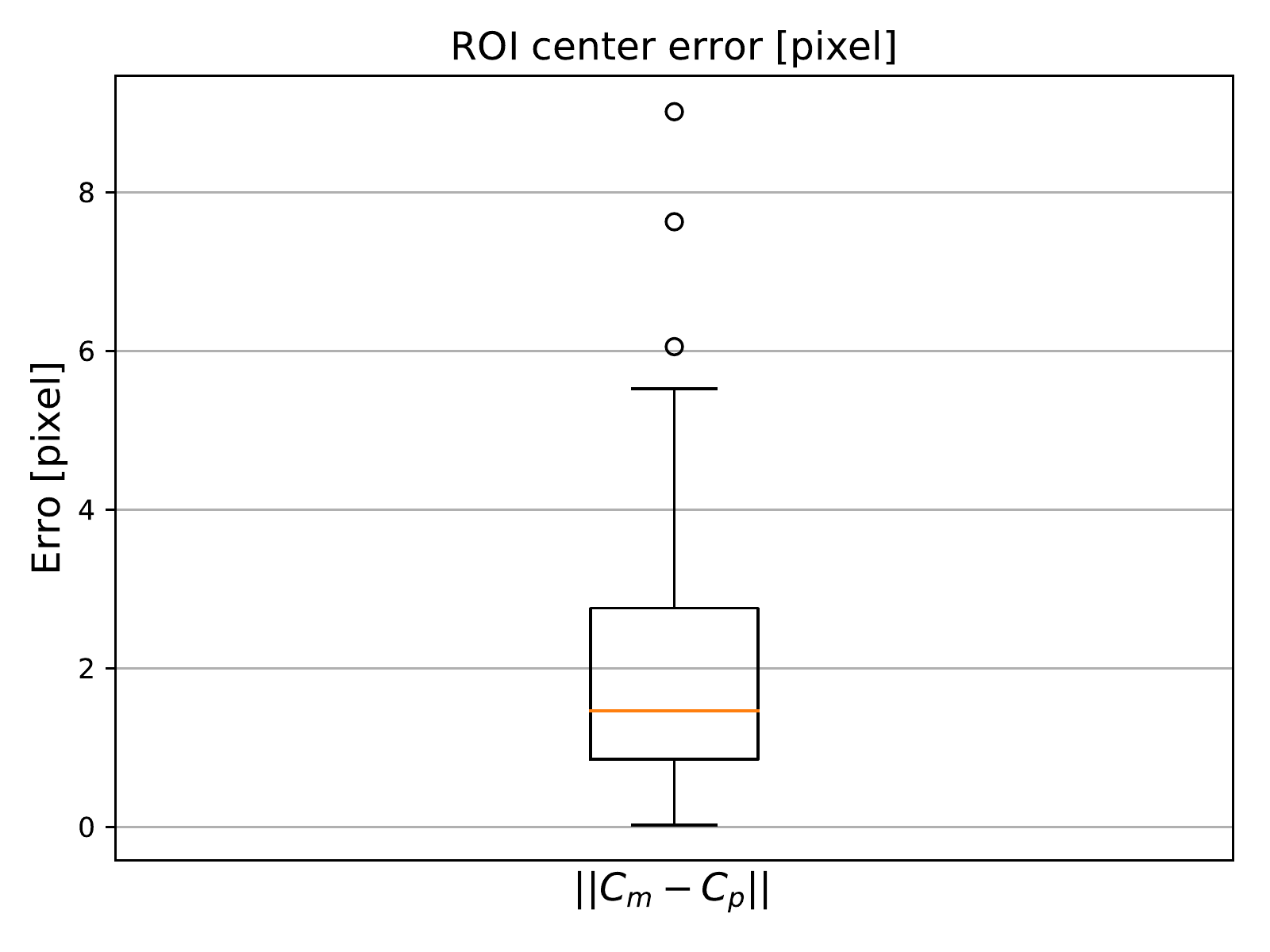}
        \caption{\label{fig_roi_error}}
    \end{subfigure}
    \caption{ROI detection accuracy. (a) 2D spatial error for detecting the center of the left ventricle. (b) Absolute error.\label{fig_roi_accuracy}}
\end{figure*}

\begin{figure*}[!ht]
\centering
\includegraphics[width=.4\textwidth]{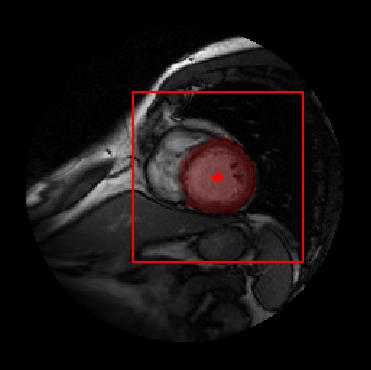}
\caption{Example of the region around the left ventricle detected by the proposed approach. The myocardial tissue and blood pool classification is depicted in red as well as the center of the left ventricle in the original image.\label{fig_roi_detection}}
\end{figure*}

The spatial difference and the mean squared error between the center of mass derived from the  manual, $C_m$, and predicted upsampling segmentation  (LV + myocardial tissue) are used to measure the CNN accuracy for detecting a ROI around the LV.
Results show that the proposed CNN approach reaches a proper accuracy for detecting a ROI around the LV with a  mean error and a spatial  difference below 2 and 4 pixels respectively (Fig.~\ref{fig_roi_error} and Fig.~\ref{fig_roi_diff}). Also, a qualitative analysis shows that the proposed approach reaches a suitable precision for this task. An example of the LV-ROI detection approach is depicted in  Fig.~\ref{fig_roi_detection}.

\subsection*{Myocardial tissue clasification}
The \Unet proposed in~\cite{Curiale2017l} was evaluated with respect to two new architectures on both datasets with a shrink factor of two (i.e. the original input size was reduced by a factor of 2). These new architectures were designed to introduce the main ideas behind Inception and Xception into the \Unet architecture. 

\begin{figure*}[!ht]
\centering
\includegraphics[width=\textwidth]{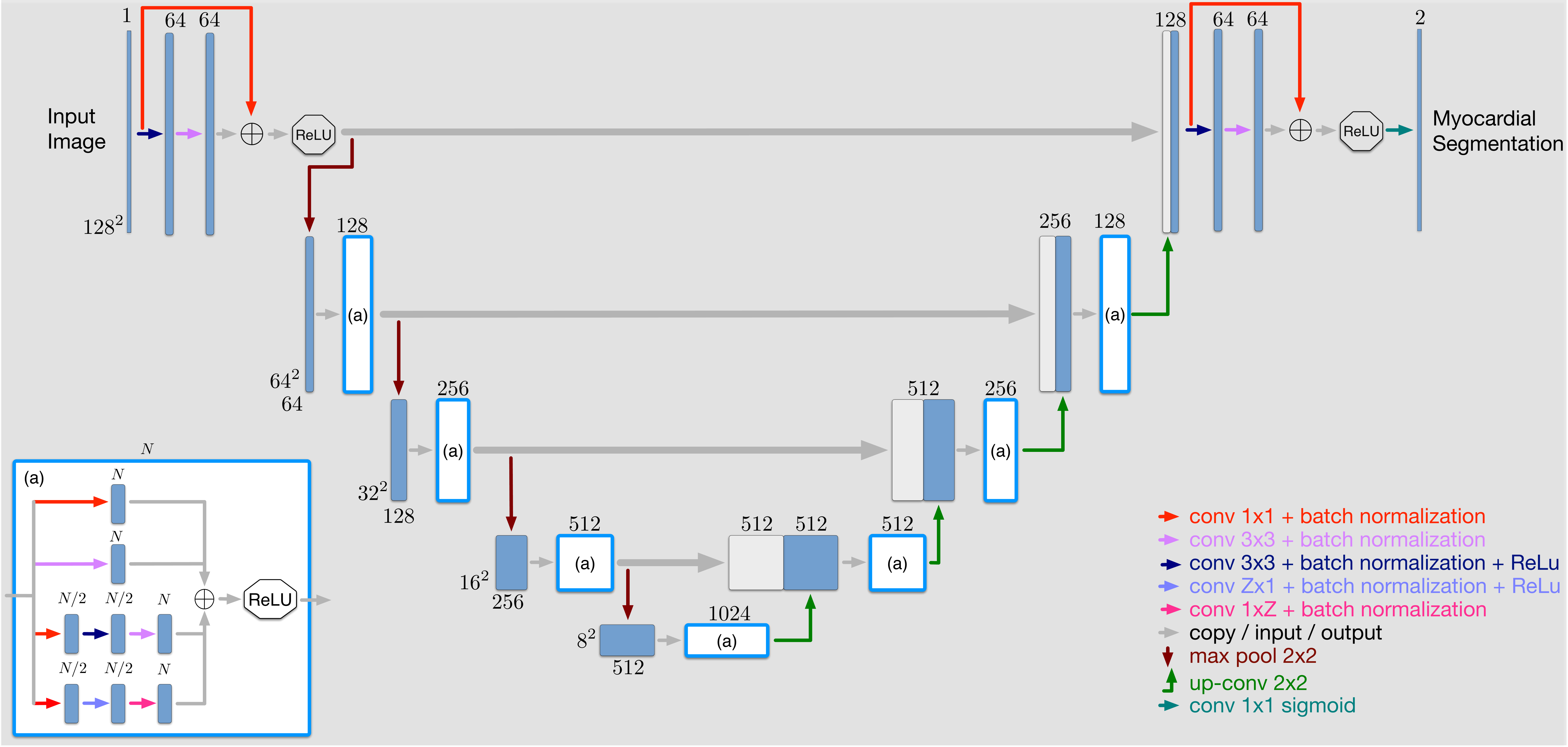}
\caption{The proposed uInception architecture for myocardial segmentation in cardiac MRI which introduces the idea of sparsity described in the Inception architecture. The number of channels/features is denoted on top of the box and the input layer dimension is provided at the lower left edge of the box. The arrows denote different operations according to the legend. Withe blocks correspond to the differences with respect to the proposed approache. The Zx1 and 1xZ convolutions refer to convolutions with different sizes according to the level, i.e. $Z = [7,6,5]$ and $[5,6,7]$ for the encoding and decoding path, and $Z=4$ for the  level 5. \label{fig_uInception}}
\end{figure*}

The sparsity of the information can be covered by convolutions over larger patches as it was pointed out in the Inception architecture. This idea is introduced into the \Unet architecture (uInception) as it is shown in Fig.~\ref{fig_uInception}, where the main differences with respected to the proposed approach are depicted as white blocks. To increase the kernel size and maintain the same network complexity it is necessary to perform a dimension reduction/expansion as it was done in the Inception architecture. In particular, this step is performed  by a 1~x~1 convolution (N/2 and N  features) without activation (see box (a) in Fig.~\ref{fig_uInception}). It is important to note that the size of some of the convolution layers changes according to the level. These convolution layers are  described  in Fig.~\ref{fig_uInception} as Z~x~1 and 1~x~Z where $Z = [7,6,5]$ and $[5,6,7]$ for the encoding and decoding path, and $Z=4$ for the  level 5.

\begin{figure*}[!t]
\centering
\includegraphics[width=\textwidth]{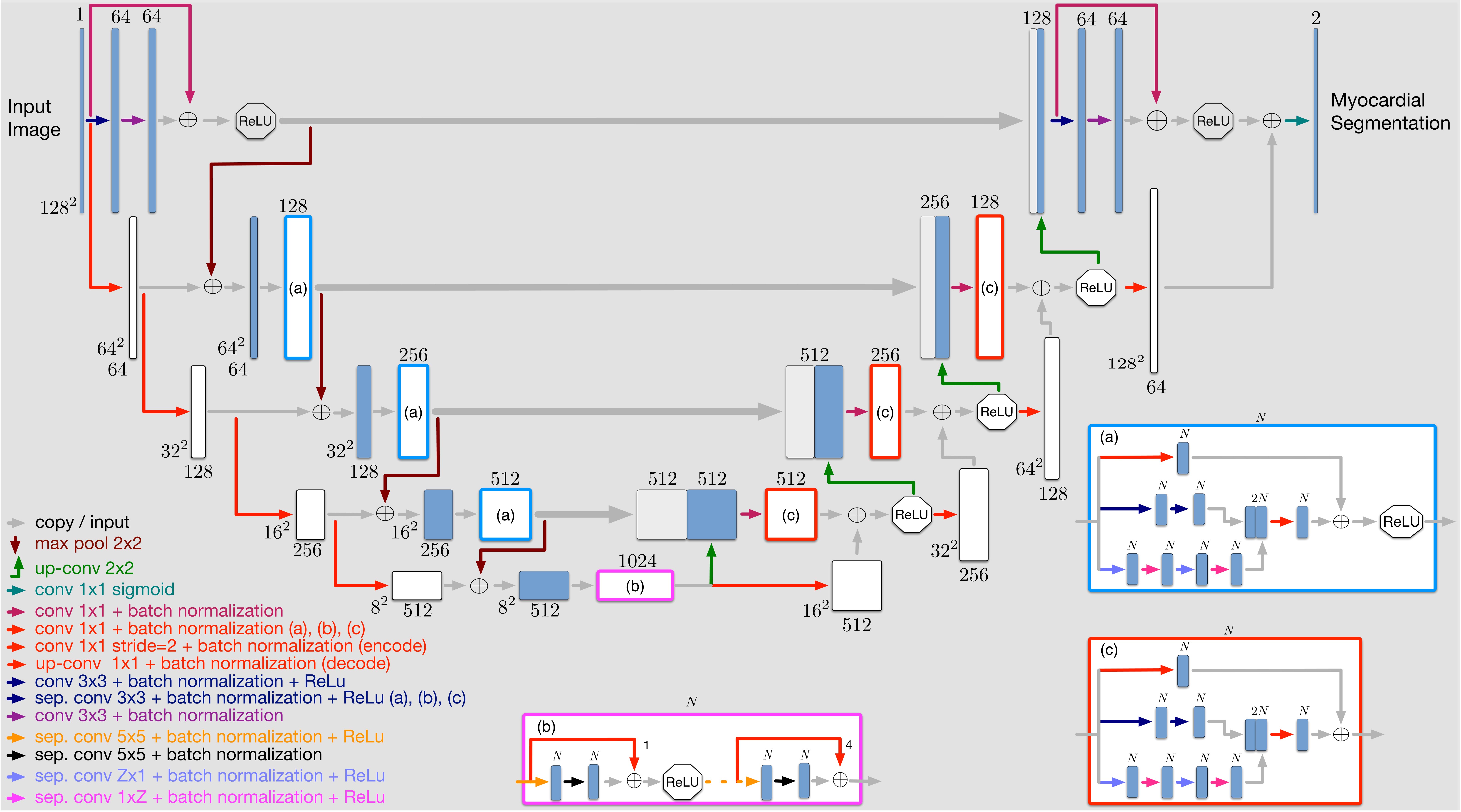}
\caption{The proposed uXception architecture  for myocardial segmentation in cardiac MRI which introduces the idea of depthwise separable convolution used in the Xception  and the sparcity described in the Inception. The number of channels/features is denoted on top of the box and the input layer dimension is provided at the lower left edge of the box. The arrows denote different operations according to the legend. Withe blocks correspond to the differences with respect to the proposed approache. The Zx1 and 1xZ convolutions refer to convolutions of different sizes according to the level, i.e. $Z = [9,7,5]$ and $Z = [5,7,9]$ for the encoding and decoding phases. \label{fig_uXception}}
\end{figure*}

The other architecture studied combines the concepts of the depthwise separable convolution described in the Xception~\citep{Chollet2017z} and  the sparsity idea of the Inception architecture. The new architecture, named uXception, is depicted in Fig.~\ref{fig_uXception}. In a similar way as it was done for the uInception, the main differences with respected to the proposed approach are depicted as white blocks.
In this case, the first level is exactly the same as the uInception and \Unet with batch normalization (BN) and residual learning (RS). Then, each level of the encode path introduce a residual input learning with a 1~x~1 convolution without activation. Also, a mirrored residual learning is introduced into the decode path by means of an 1~x~1 up-convolution (see the red arrow after the ReLu activation for the decode path in Fig.~\ref{fig_uXception}).
Finally, the last level consists of four blocks of two 5~x~5 separable convolution, followed each convolution with a batch normalization and summed up with a residual learning (Fig.~\ref{fig_uXception} box (b)). Furthermore,  the size of  some of the  convolution layers changes according to the level in the same way as it was described in the uInception architecture.

\begin{figure*}[!t]
\centering
\includegraphics[width=.49\textwidth]{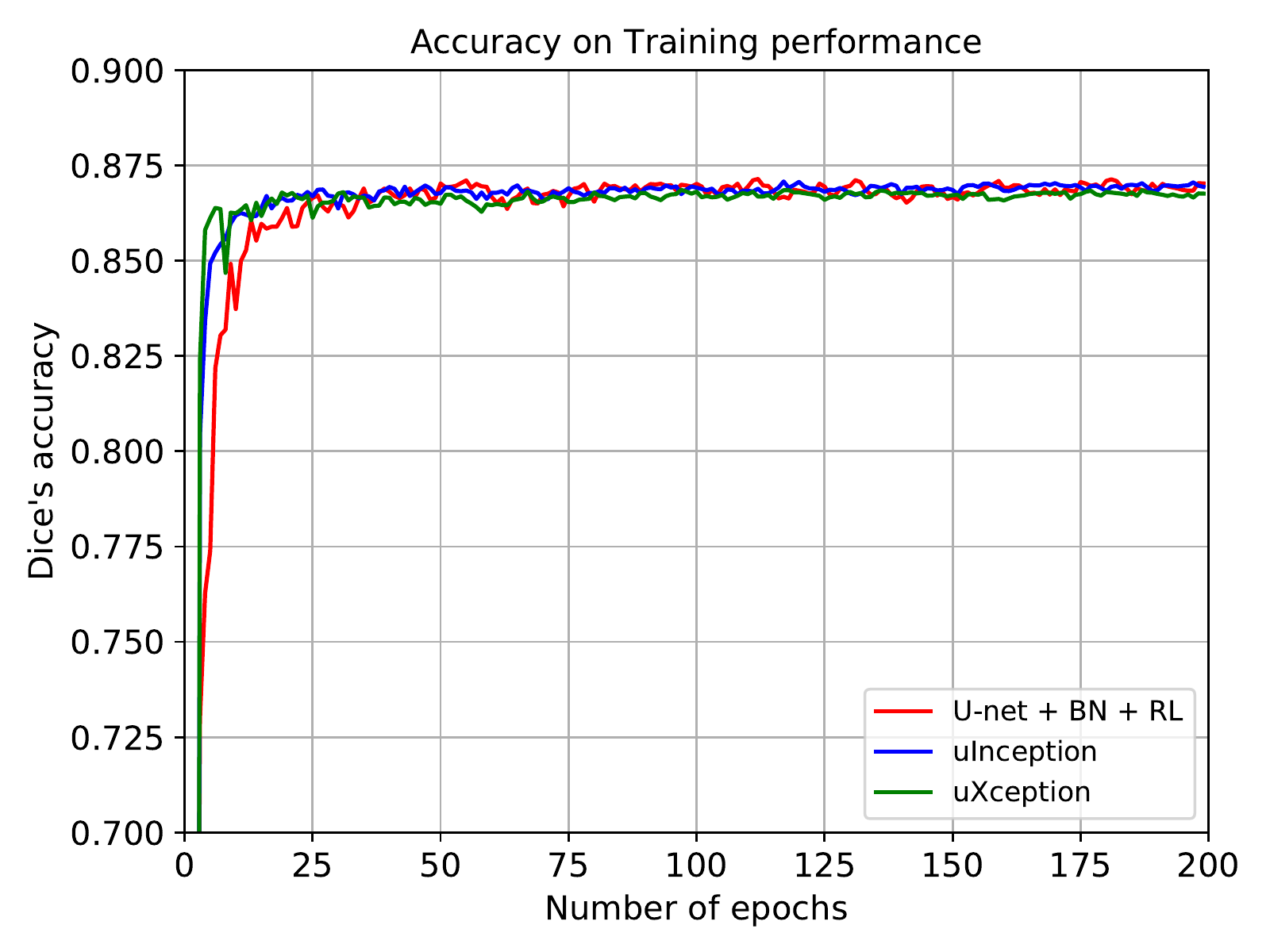}
\caption{Mean accuracy over the 5 fold cross-validation for the architectures studied with an input shrink factor of 2 (i.e. the original input size was reduced by a factor of 2).\label{fig_architectures_analysis_all}}
\end{figure*}

Empirical results show that the uInception and uXception outperform the \Unet accuracy on the first 25 epochs,  and they achieve similar performance after 200 epochs (Fig.~\ref{fig_architectures_analysis_all} and Table~\ref{table_accuracy}). 
We believe that this similar behavior is due to the reduced number of images used for training.
Indeed, if the size of the dataset increases, it is expected that the uInception and uXception outperform the \Unet with BN and RL because they present differences in the network capacity. Also, it is important to note that the uXception approach reduces the network complexity in about 25.5\% with respect to the number of parameters to be trained (Table~\ref{table_accuracy} parameter count), and  it keeps the same accuracy for myocardial tissue classification. 
In the same way as the Xception improves the accuracy of the Inception when the dataset size increase~\citep{Chollet2017z}, we expect that the uXception outperforms the uInception architecture too.

\begin{table}[t]
\begin{tabular}{l|cccc}
\hline
Architecture 				& Parameter count & Dice's (std) & MSE (std) & MAE (std) \\
\hline
   	\Unet - BN - RL  		& 32,455,682 & 0.870 (0.0053)	& 0.0135 (0.0006)	& 0.0137 (0.0006)\\
   	uInception  			& 35,858,242 &0.869 (0.0051)	& 0.0136 (0.0008)	& 0.0138 (0.0008)\\
	uXception  			& 24,181,570 &0.868 (0.0047) 	& 0.0138 (0.0007) 	& 0.0140 (0.0007)\\	
\hline
\end{tabular}
\caption{Network complexity and myocardial segmentation accuracy over the 5-fold cross-validation (mean and std) for the architectures studied with an input shrink factor of 2 (i.e. the original input size was reduced by a factor of 2). MSE: mean squared error. MAE: mean absolute error. BN: Batch normalization. RL: Residual learning  {\color{Red} revisar...}. \label{table_accuracy}}
\end{table}

\begin{figure*}[!t]
\centering
\begin{subfigure}[b]{0.48\textwidth}
\includegraphics[width=\textwidth]{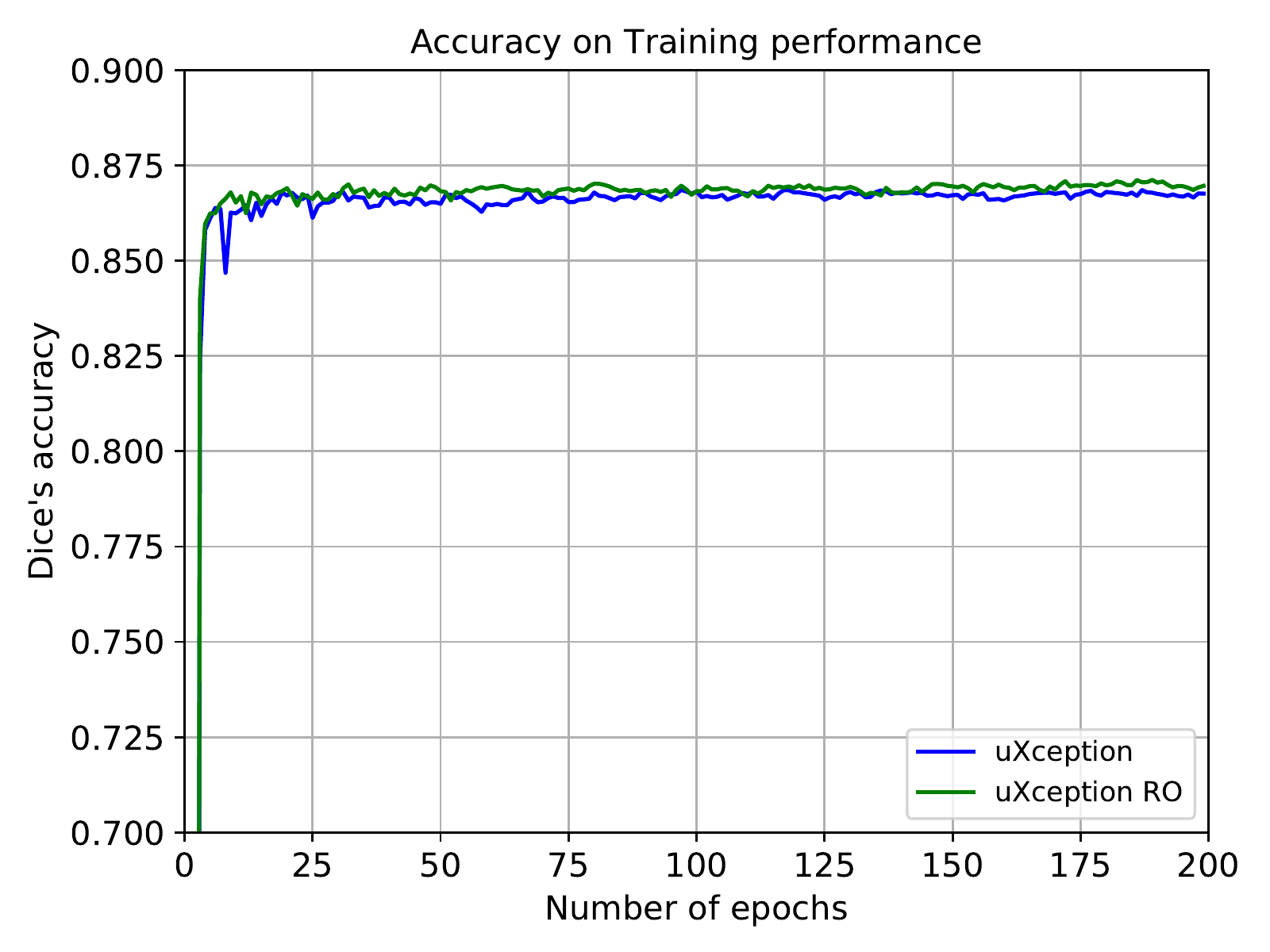}
\caption{\label{fig_architectures_analysis_uXception}}
\end{subfigure}
~ 
\begin{subfigure}[b]{0.48\textwidth}
\includegraphics[width=\textwidth]{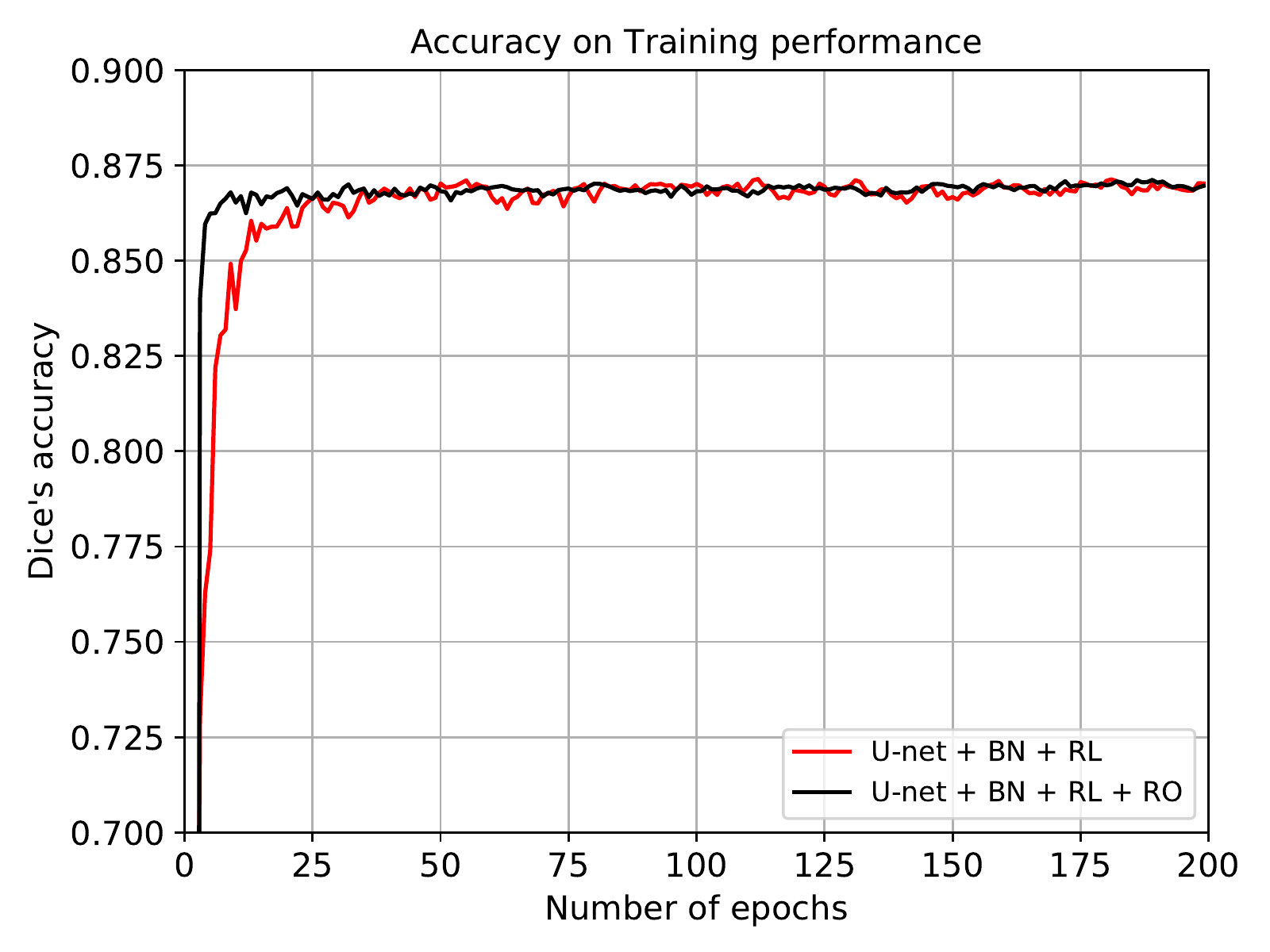}
\caption{\label{fig_architectures_analysis_unet}}
\end{subfigure}
\caption{Mean accuracy over the 5 fold cross-validation for the uXception and \Unet architectures  with an input shrink factor of 2 (i.e. the original input size was reduced by a factor of 2). RO: residual output. \label{fig_ro_analysis}}
\end{figure*}

The residual learning from level to level used in the Xception architecture was introduced into the uXception approach as the residual input and output learning (see Fig.~\ref{fig_uXception} red arrows in the encode and decode path). Both residual reinforcement, input and output (RO), were evaluated first into the uXception architecture, and then into the \Unet. Results show that the residual input can be removed from the uXception approach without losing accuracy. Instead, it was observed a small improvement on the myocardial tissue classification if it is removed (Fig.~\ref{fig_architectures_analysis_uXception}).  Then, the effect of the residual output learning  was analyzed for the \Unet architectures. The analysis of the residual input learning was omitted because it showed to degrade the myocardial tissue classification for the uXception architecture. 
Figure~\ref{fig_architectures_analysis_unet} shows that the use of RO improves the tissue classification in an early stage of training for the \Unet architecture. Due to the reduced number of patient used in this study, both architectures get similar accuracy after 200 epochs. But, we believe that  \Unet with RO will outperform the \Unet architecture when the dataset size increases because the network capacity increases when the RO is introduced, in a similar way as what happened in~\cite{Chollet2017z} with the Xception and Inception architectures.

\begin{figure*}[!t]
    \centering
        \includegraphics[width=.5\textwidth]{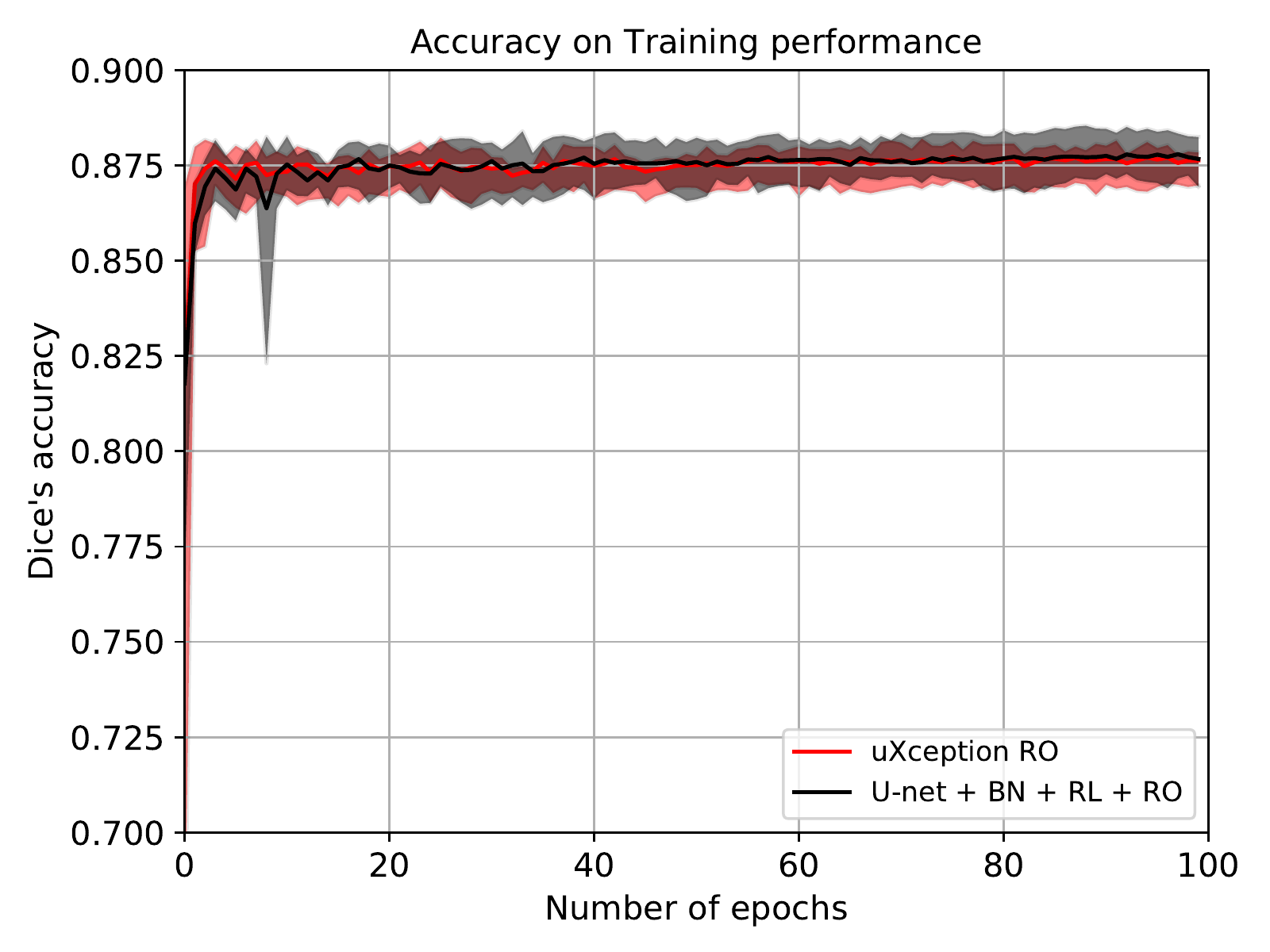}
    \caption{Accuracy on training performance over the 5 fold cross-validation for the proposed \Unet and uXception architecture with residual output for myocardial  segmentation in cardiac MRI.\label{fig_unet_ro_uXception_osize}}
\end{figure*}

\begin{figure*}[!t]
    \centering
        \includegraphics[width=.48\textwidth]{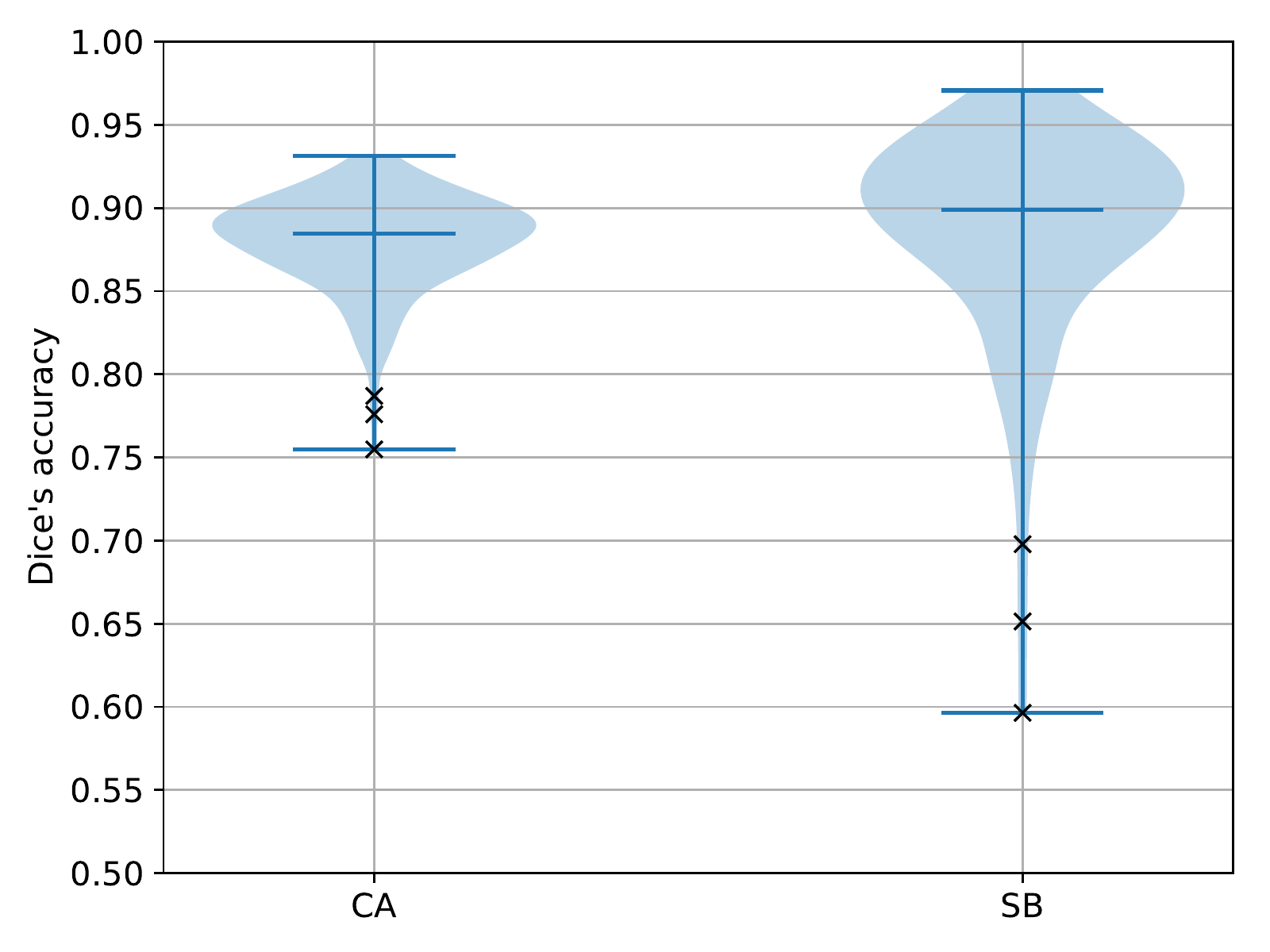}
        \includegraphics[width=.48\textwidth]{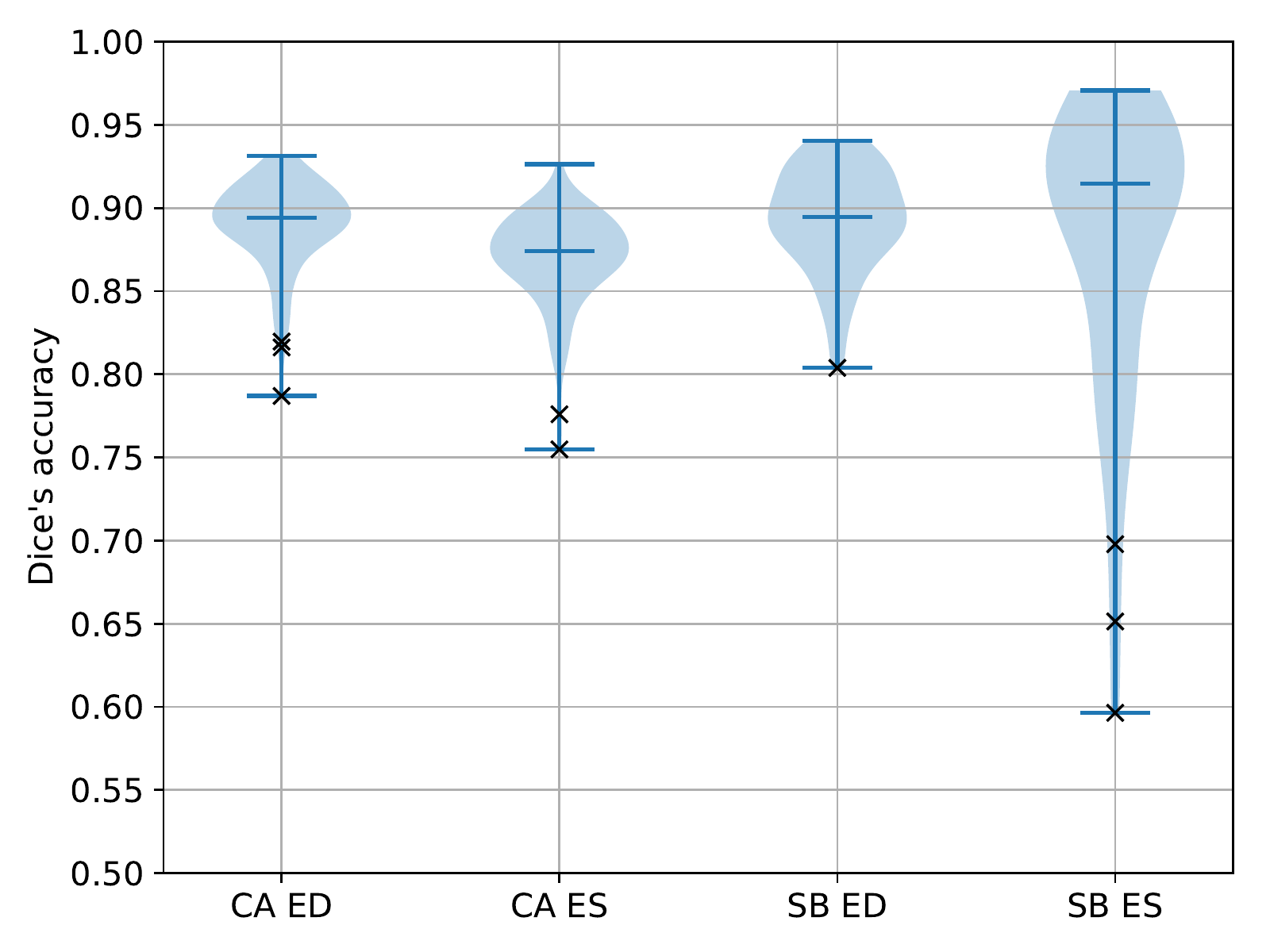}
    \caption{ Accuracy of the proposed architecture for myocardial  segmentation in cardiac MRI for each dataset (left), and also for end-diastole and end-systole (right). Only the endocardial contours were used for the SB database at end-systole. CA: Cardiac Atlas dataset; SB: Sunnybrooks dataset; ED: end-diastole; ES: end-systole.\label{fig_proposed_accuracy}}
\end{figure*}

Last, the proposed \Unet architecture with RO and the uXception were evaluated on both dataset without using a shrinking factor. Figure~\ref{fig_unet_ro_uXception_osize} shows that the \Unet with RO architecture is slightly more accurate than uXception for the datasets used. And, the \Unet with RO approach reaches a suitable accuracy in few training iterations.  i.e. in 100 epochs. In this case, the proposed CCN reaches a Dice's median value of $\sim0.9$ for both dataset (Fig.~\ref{fig_proposed_accuracy}). However, it can be seen that the precision tends to be degraded at end-systole (Fig.~\ref{fig_proposed_accuracy} right plot). This degradation happens, among other things,  because  the myocardial tissue at end-diastole is generally more compact, and also, there is more end-diastole images than end-systole. In fact, the Sunnybrooks dataset  provides the myocardial contours in all slices at ED, and only the endocardial contours at ED.

\begin{figure*}[!t]
    \centering
    \begin{subfigure}[b]{0.635\textwidth}
        \includegraphics[width=\textwidth]{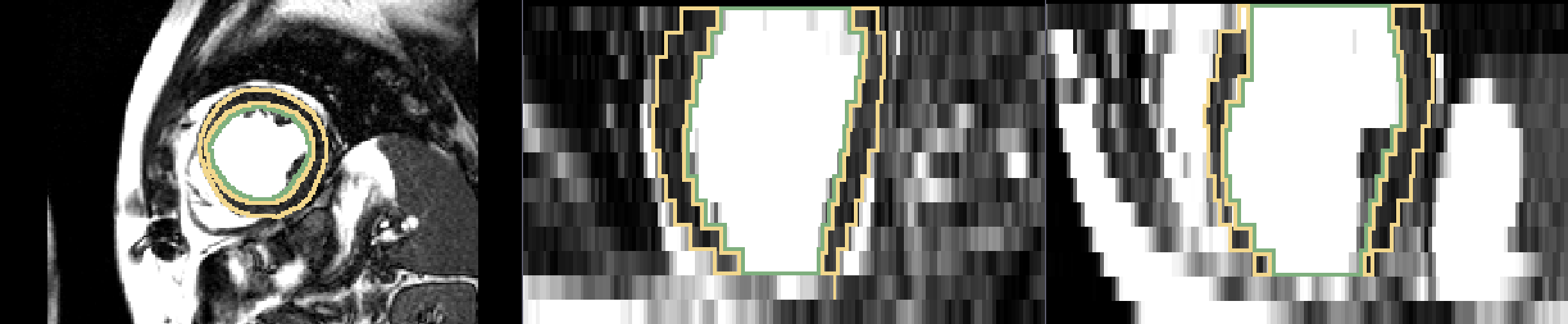}
        \includegraphics[width=\textwidth]{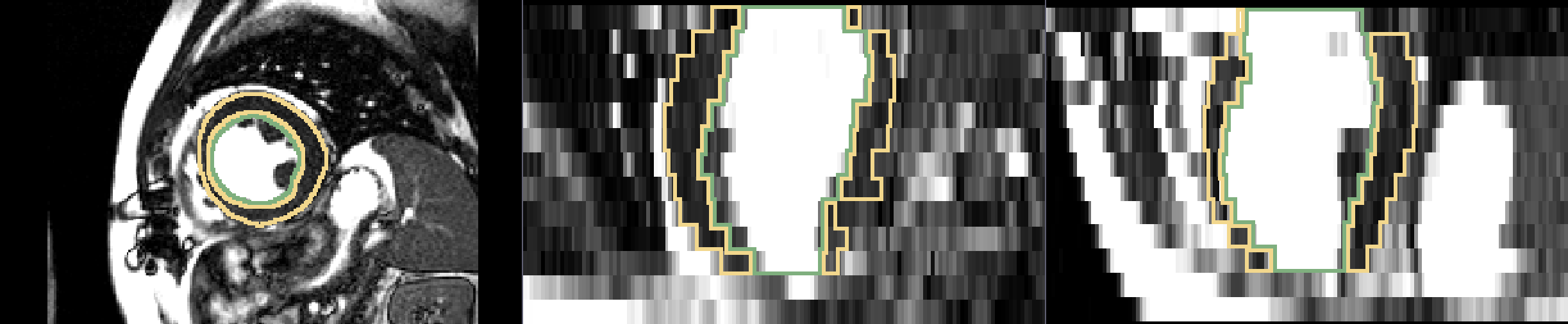}
        \caption{\label{fig_myo_segmentation_2d}}
    \end{subfigure}
    \begin{subfigure}[b]{0.355\textwidth}
        \includegraphics[width=\textwidth]{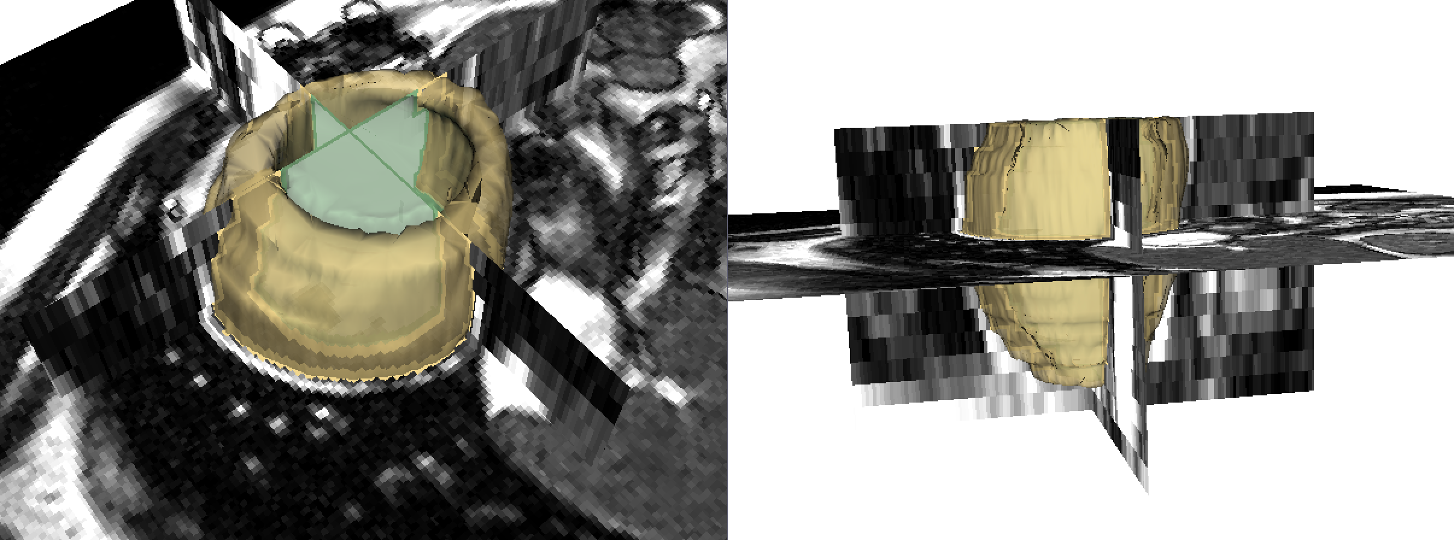}
        \includegraphics[width=\textwidth]{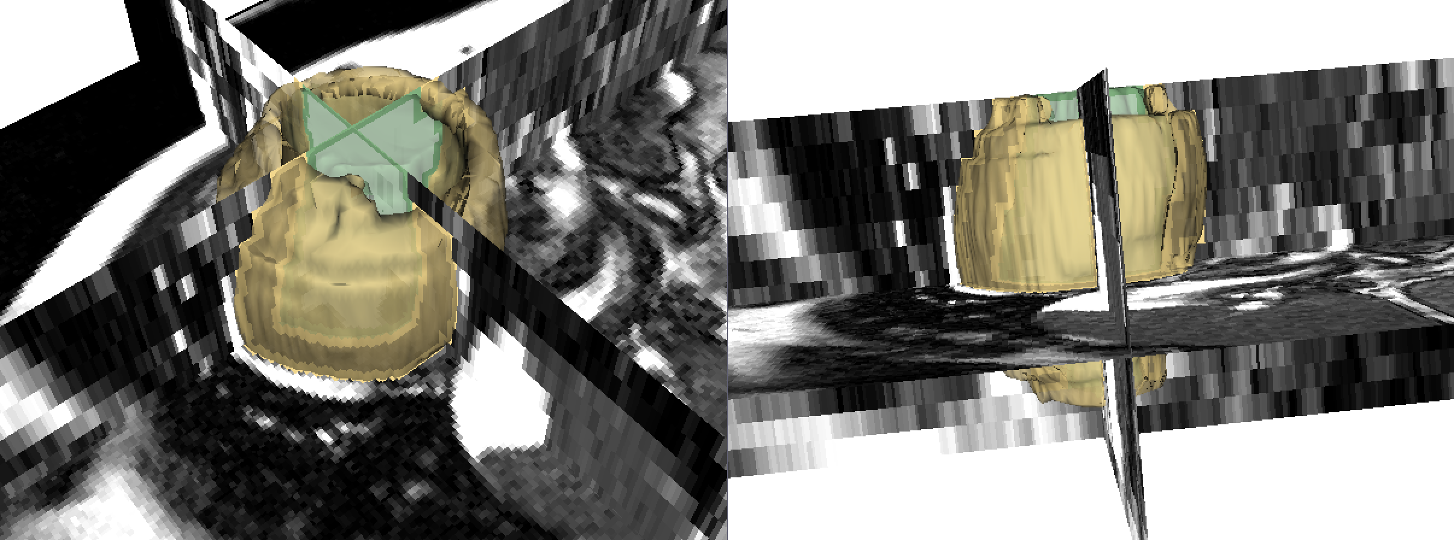}
        \caption{\label{fig_myo_segmentation_3d}}
    \end{subfigure}
    \caption{Examples of the proposed method for myocardial tissue and endocardial (blood pool) classification on a patient of the CA dataset for end-diastole (up-row) and end-systole (low-row). (a) The proposed myocardial segmentation is presented in three orthogonal views. (b) 3D view of the proposed myocardial segmentation. \label{fig_myo_segmentation}}
\end{figure*}

\begin{figure*}[t!]
\centering
\includegraphics[width=.95\textwidth]{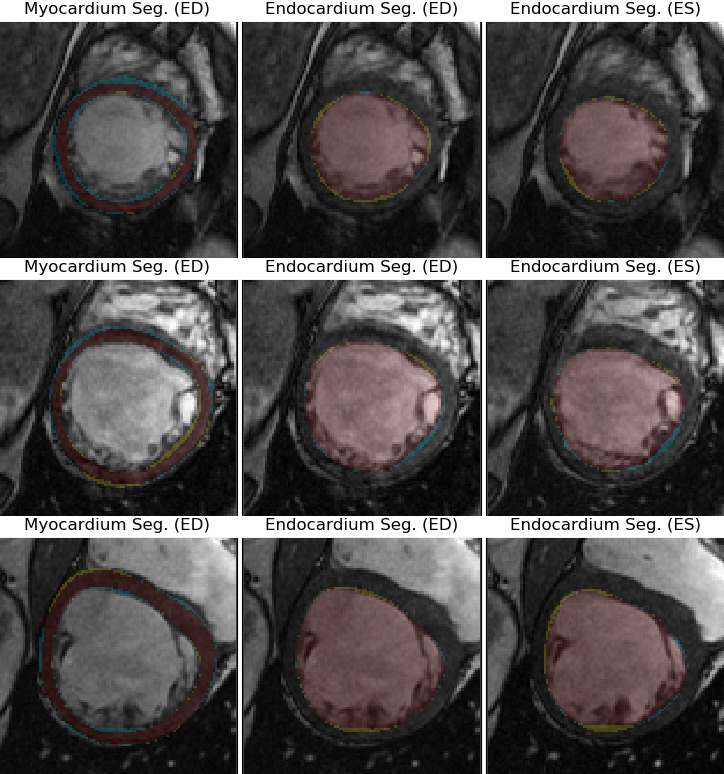}
\caption{Qualitative results for a patient in the SB dataset. Three different short axis slices are plotted from apex (upper) to base (down) of the left ventricle for the myocardial tissue and the endocardium (blood pool) at end-diastole and only the endocardium at end-systole.  In brown it depicts those voxels where the manual and the proposed automatic classification overlaps, in dark yellow and cian  are depicted those pixels corresponding to the manual and automated classification without overlapping respectively. \label{fig_final_unet_quantitative}}
\end{figure*}

A 3D examples of the proposed fully automatic  myocardial tissue segmentation on the CA dataset can be seen in Fig.~\ref{fig_myo_segmentation}.  Figure~\ref{fig_myo_segmentation_2d} shows the myocardial segmentation in three orthogonal views and Fig.~\ref{fig_myo_segmentation_3d}  shows the segmentation in two 3D views.
Qualitative details of the proposed approach for myocardial segmentation for a patient on the Sunnybrooks dataset can be seen in several slices in Fig.~\ref{fig_final_unet_quantitative}. They show that the proposed myocardial segmentation provides a suitable approach for myocardial segmentation in cardiac MRI.  In brown it depicts those voxels where the manual myocardial segmentation  and the proposed automatic segmentation overlaps, in dark yellow and cian  are depicted those pixels corresponding to the GT and the automated segmentation without overlapping respectively. As it can be seen only a reduced number of voxels correspond to a non-overlapping classification, especially for the LV ventricle.

\subsection*{Physiological validation}
Six physiological measures have been derived from the myocardial tissue classification: end-diastolic volume (EDV),  end-systolic volume (ESV),  SV,  LVM at ED,  CO and  EF. As it was described in the introduction, these measures are one of the most relevant global structural features.  In particular, the  ESV measures for three patients  were identified as  outliers and they will be discussed separately in the next section.

Table~\ref{table_phys_val} summarizes the errors and the Pearson's correlation coefficient of the six physiological measures derived from the  proposed method with  respect to those derived from the experts. Whereas in Table~\ref{table_phys_val2} a bibliographical comparison is performed. In general, our measures are comparable to those reported by other authors.

\begin{table}[!h]
\center
\begin{tabular}{l|ccccc}
\hline
Parameters & Mean error & Std error 	& Mean abs error& Max. abs error & $\rho$ \\
\hline
EDV [cm$^3$] 		& 0.14  & 9.51  &  7.31 & 26.55  & 0.99\\
ESV [cm$^3$] 		& 1.37  & 9.15  & 7.12  & 25.42   & 0.99\\
SV [cm$^3$] 		& -1.23   & 8.38  & 6.82  & 21.54   & 0.93\\
EF [\%] 			& -0.71   & 4.76  & 3.73 & 15.84 & 0.95\\
LVM [g] 			& -3.68   & 13.66 & 10.67 & 64.05 & 0.97\\
CO [L/min] 		& -0.03   & 0.47  & 0.32  & 1.29 & 0.93\\
\hline
\end{tabular}
\caption{Errors of the physiological measures studied for the proposed approach.\label{table_phys_val}}
\end{table}
\begin{table}[!h]
{\footnotesize
\begin{tabular}{p{2cm}|cccc|p{3.6cm}p{3.3cm}}
\cline{2-5}
& \multicolumn{4}{|c|}{Mean $\pm$ std. errors} & &\\
\hline
Reference & EDV [cm$^3$] & ESV [cm$^3$] & EF [\%] & LVM [g] & Methodology & Comments\\
\hline
\cite{Corsi2006f}& $1.0 \pm 5.0$ & $-3.0 \pm 7.0$ & $2.0 \pm 5.5$&  ---  & active contours  & Nine patients\\

\cite{Pednekar2006y}& $-1.8 \pm 11.7$ & $-10.8 \pm 8.2$ & $7.6 \pm 5.6$& --- & Fuzzy objects, Hough Transform and  minimum cost   & Six patients and eight healthy subjects\\

\cite{Geuns2006s}& $-8.1 \pm 11.5$ & $-5.9 \pm 6.3$ & $1.6 \pm 3.5$&  $7.2 \pm 15.0$  & Fuzzy objects, smooth convex hull and radial minimum cost & Seven patients and three healthy subjects\\

\cite{Mazonakis2010b}& $-6.1 \pm 7.2$ & $-3.0 \pm 5.2$ & $0.6 \pm 4.3$& $6.2\pm12.2$  & Bayesian flooding and weighted least-squares  & 18 patients; semiautomatic corrections\\

\cite{Cordero-Grande2011m}&  $-3.6 \pm 8.2$ & $-3.3 \pm 7.2$ & $1.5 \pm 3.3$&  $8.2 \pm 11.6$  & MRF based deformable model &43 patients affected by an AMI\\

\cite{Marino2016x} & $1.0 \pm 22$ & $9 \pm 23$ & $7 \pm 11$&  ---  & Maximum likelihood based on active contours & 35 patients and 15 healthy subjects \\

ours &  $0.14 \pm 9.5$ & $1.4 \pm 9.1$ & $-0.7 \pm 4.8$&  $-3.7 \pm 13.7$  & Deep learning & 137 patients\\
\hline
\end{tabular}
}
\caption{Errors (mean $\pm$ std.) in the measurement of the physiological parameters studied. Comparison with other approaches.  \label{table_phys_val2}}
\end{table}

%

\begin{figure*}[!t]
\centering
\includegraphics[width=.49\textwidth]{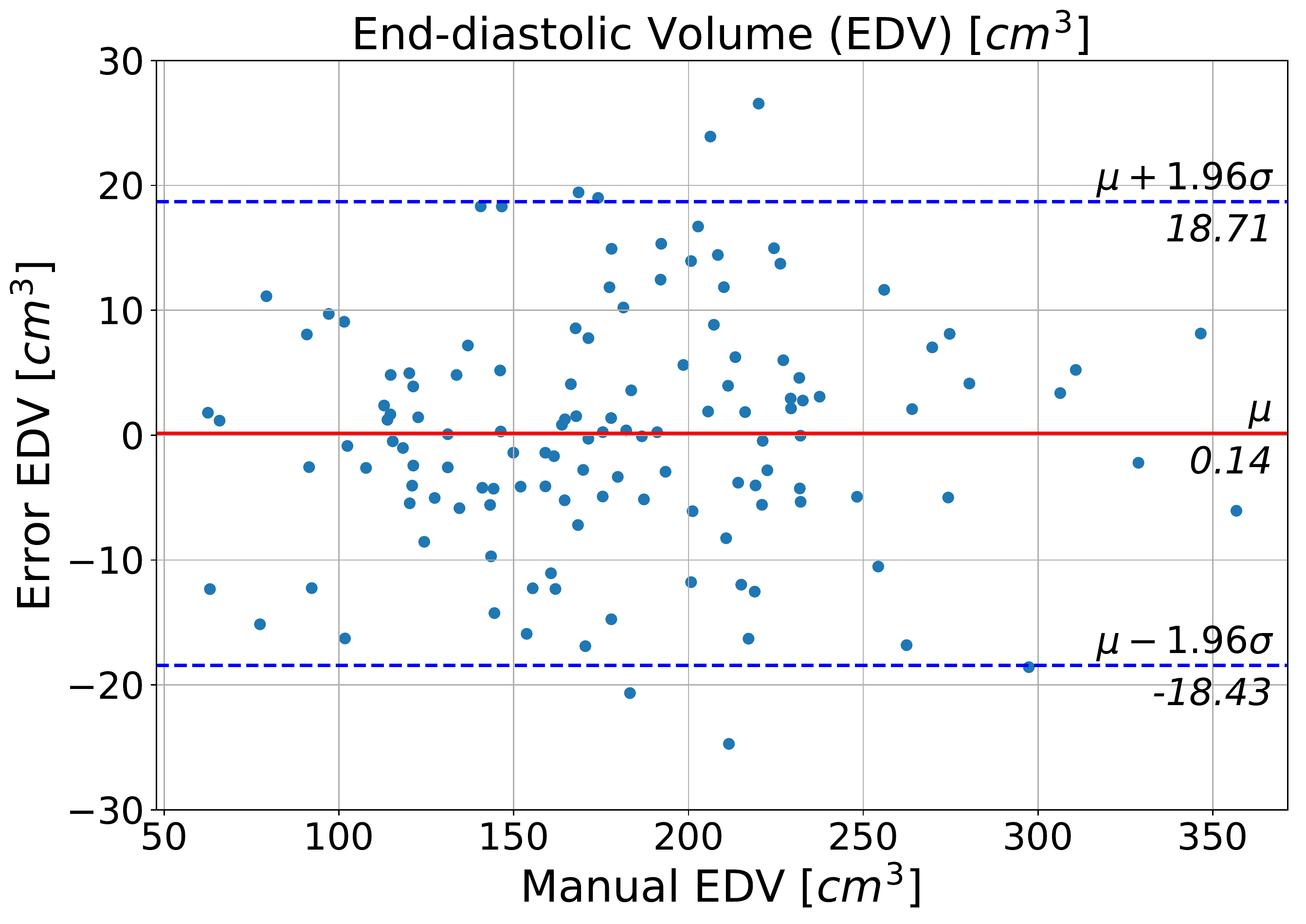}
\includegraphics[width=.49\textwidth]{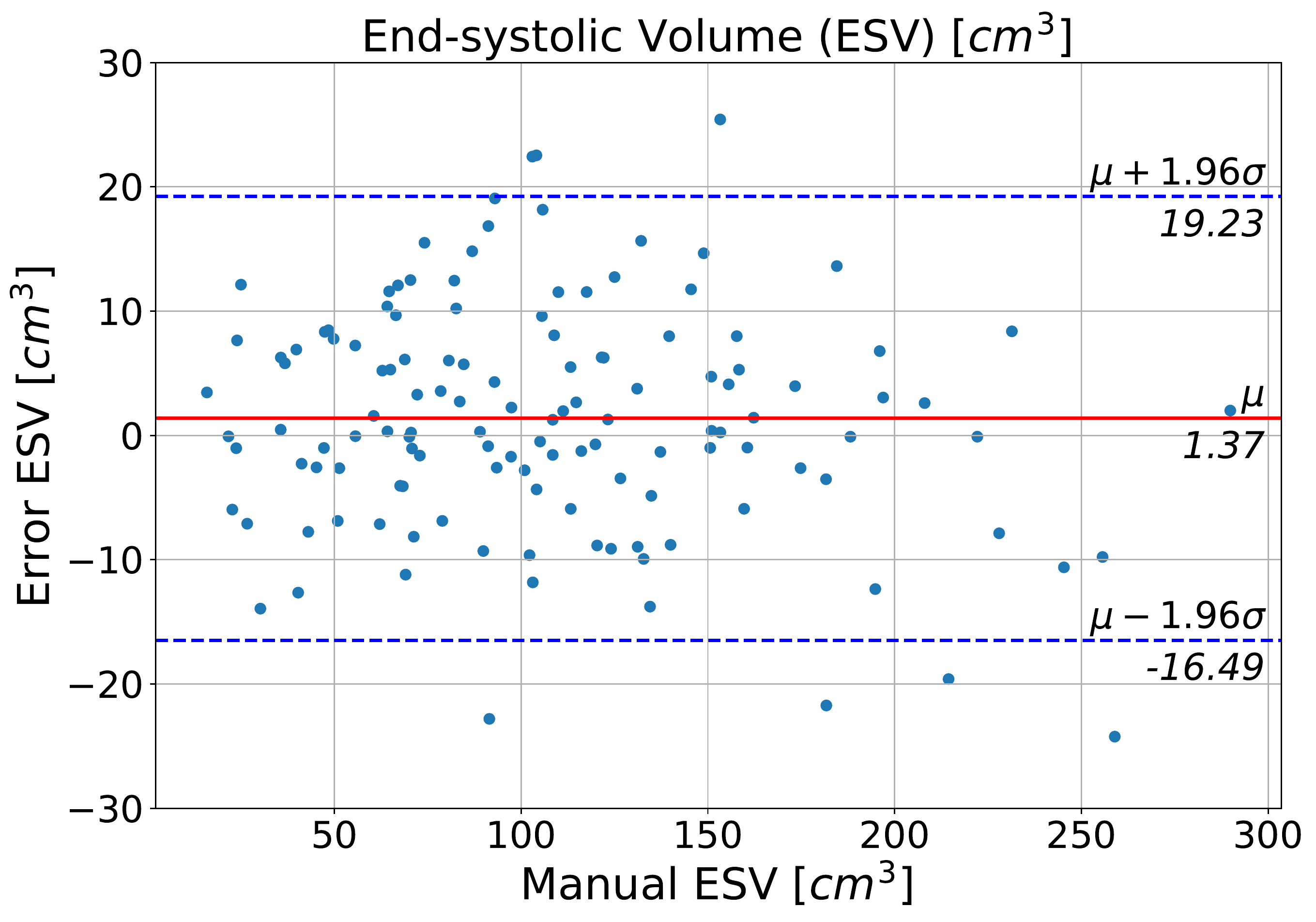}
\includegraphics[width=.49\textwidth]{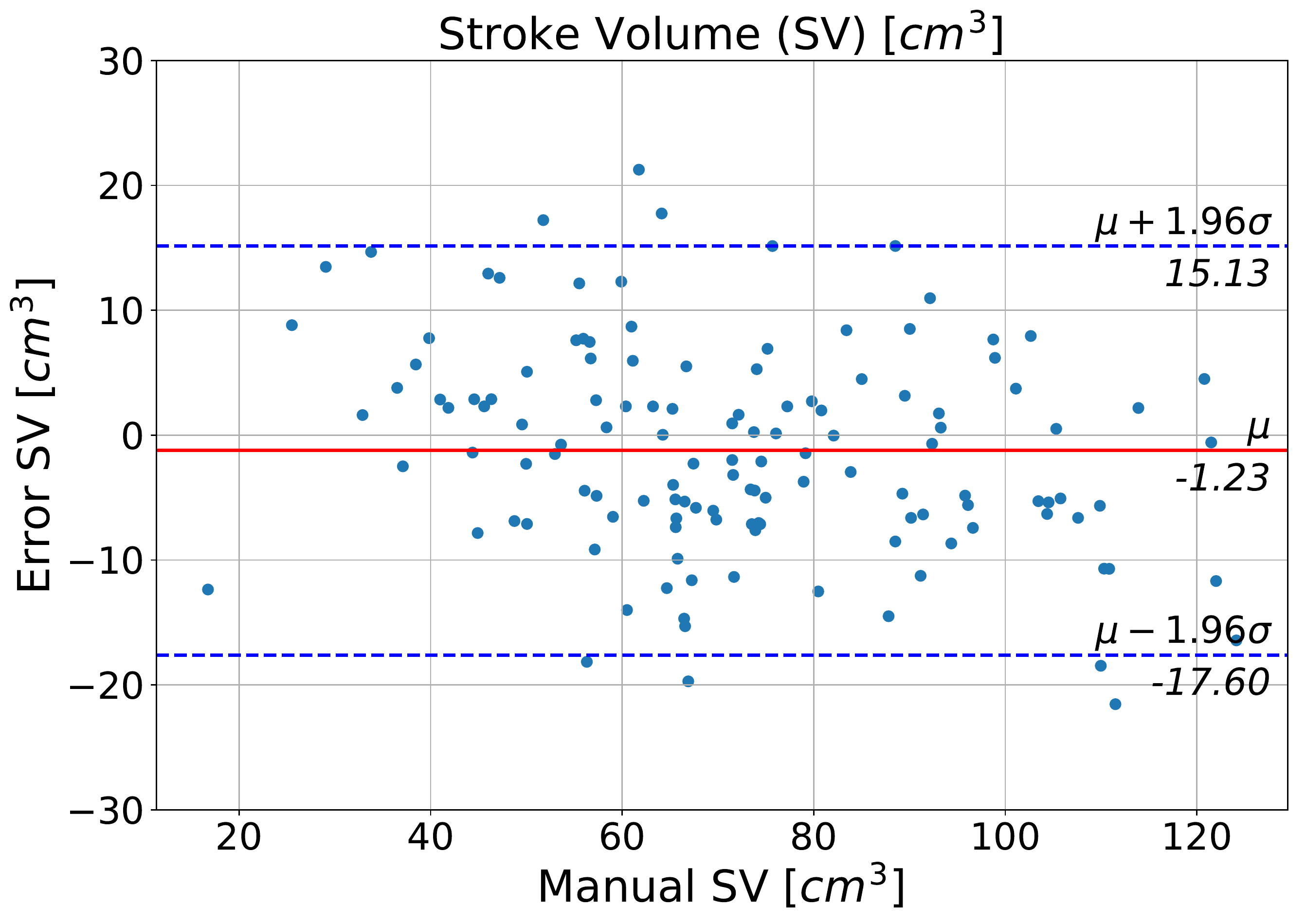}
\includegraphics[width=.49\textwidth]{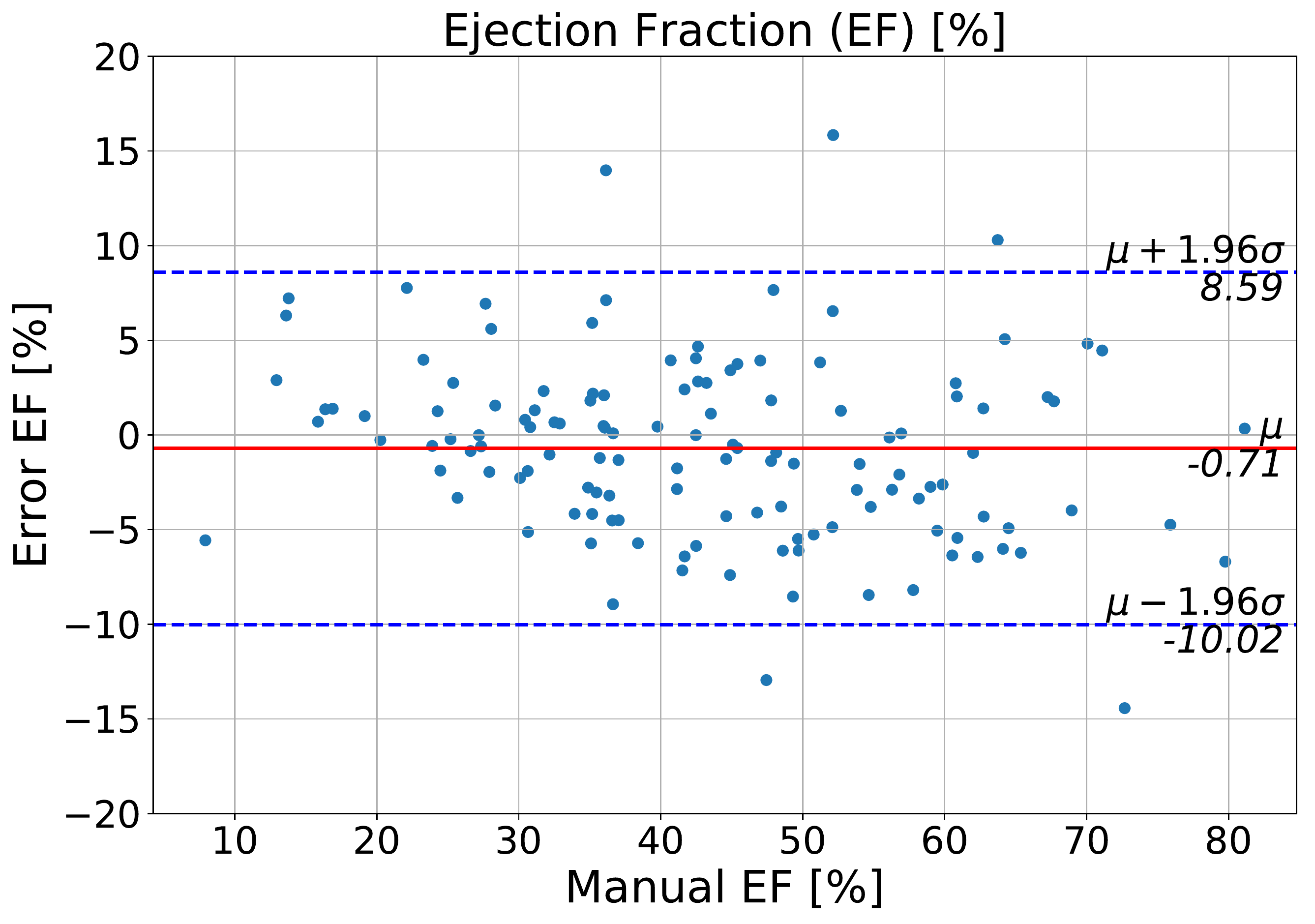}
\includegraphics[width=.49\textwidth]{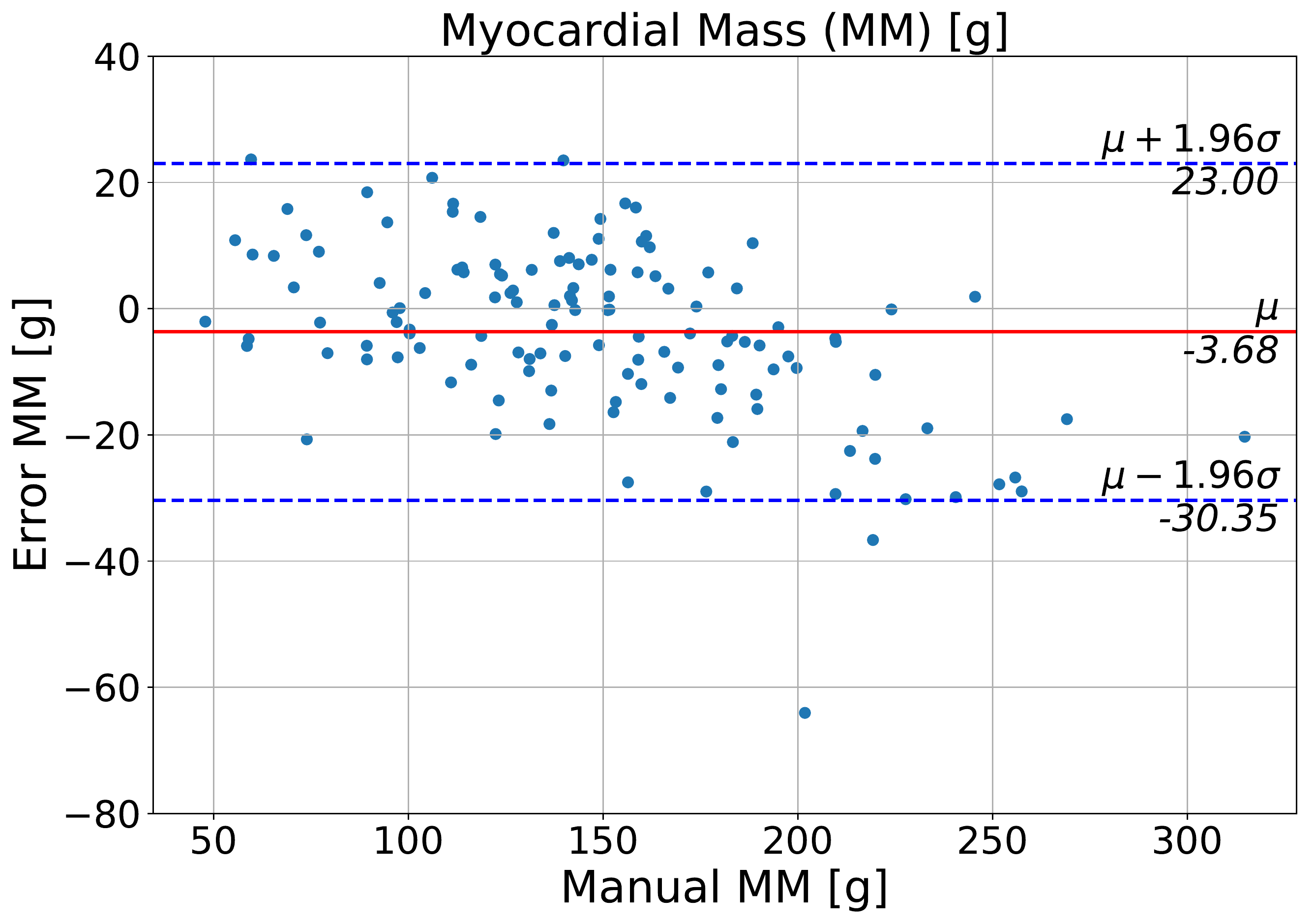}
\includegraphics[width=.49\textwidth]{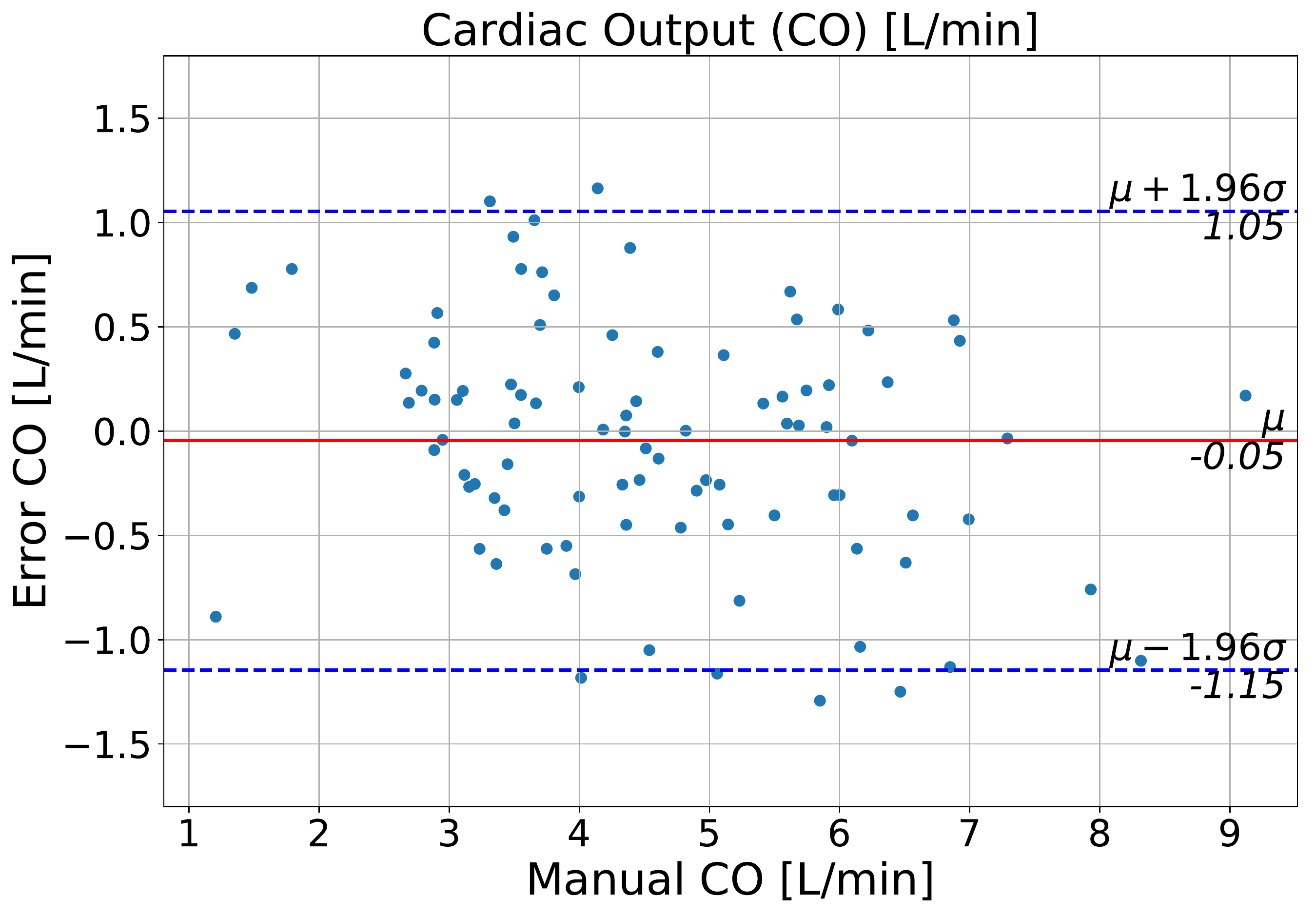}

\caption{Bland-Altman plots of the physiological measures studied. \label{fig_blandaltman}}
\end{figure*}

\begin{figure*}[!t]
\centering
\includegraphics[width=.315\textwidth]{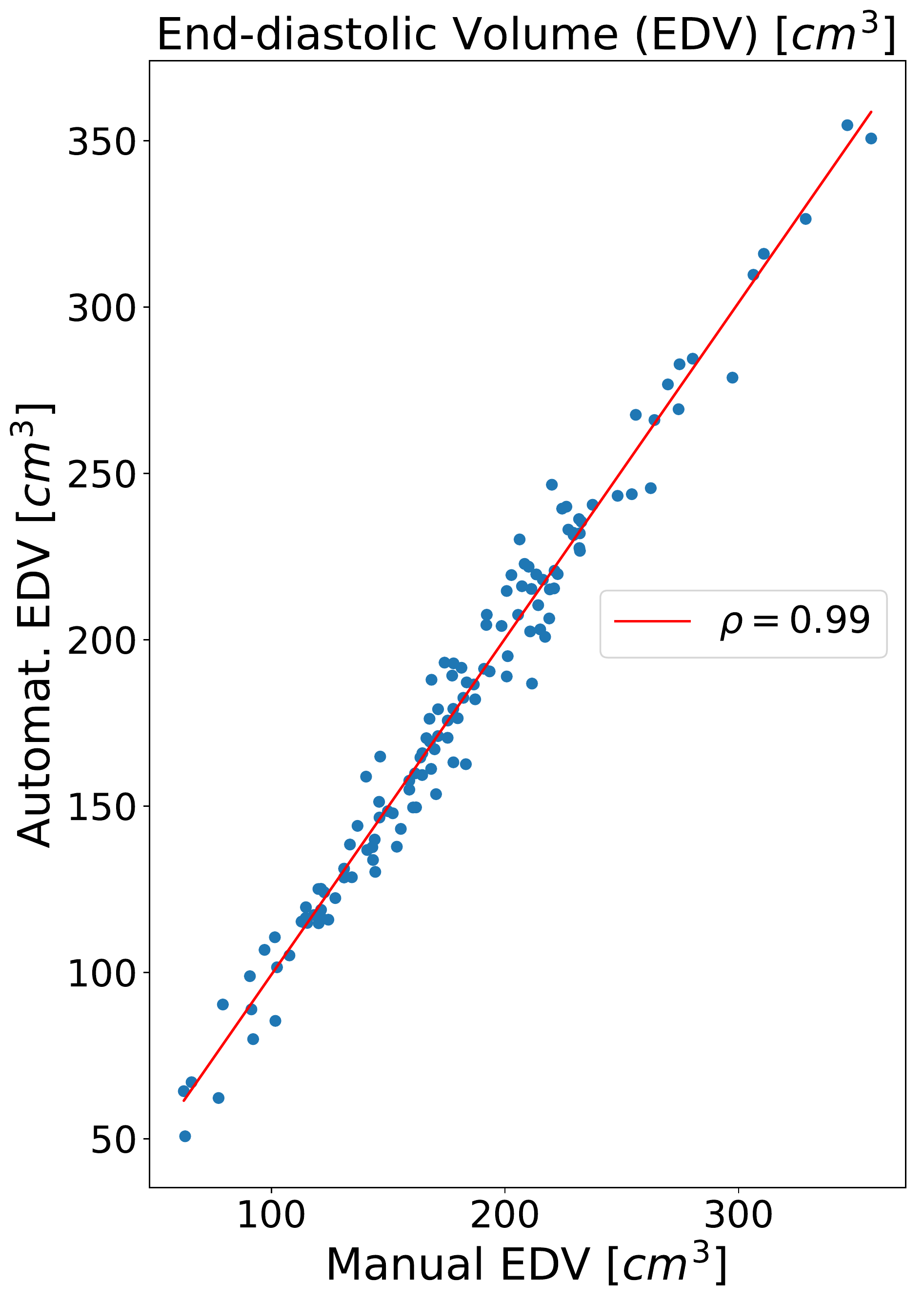}
\includegraphics[width=.32\textwidth]{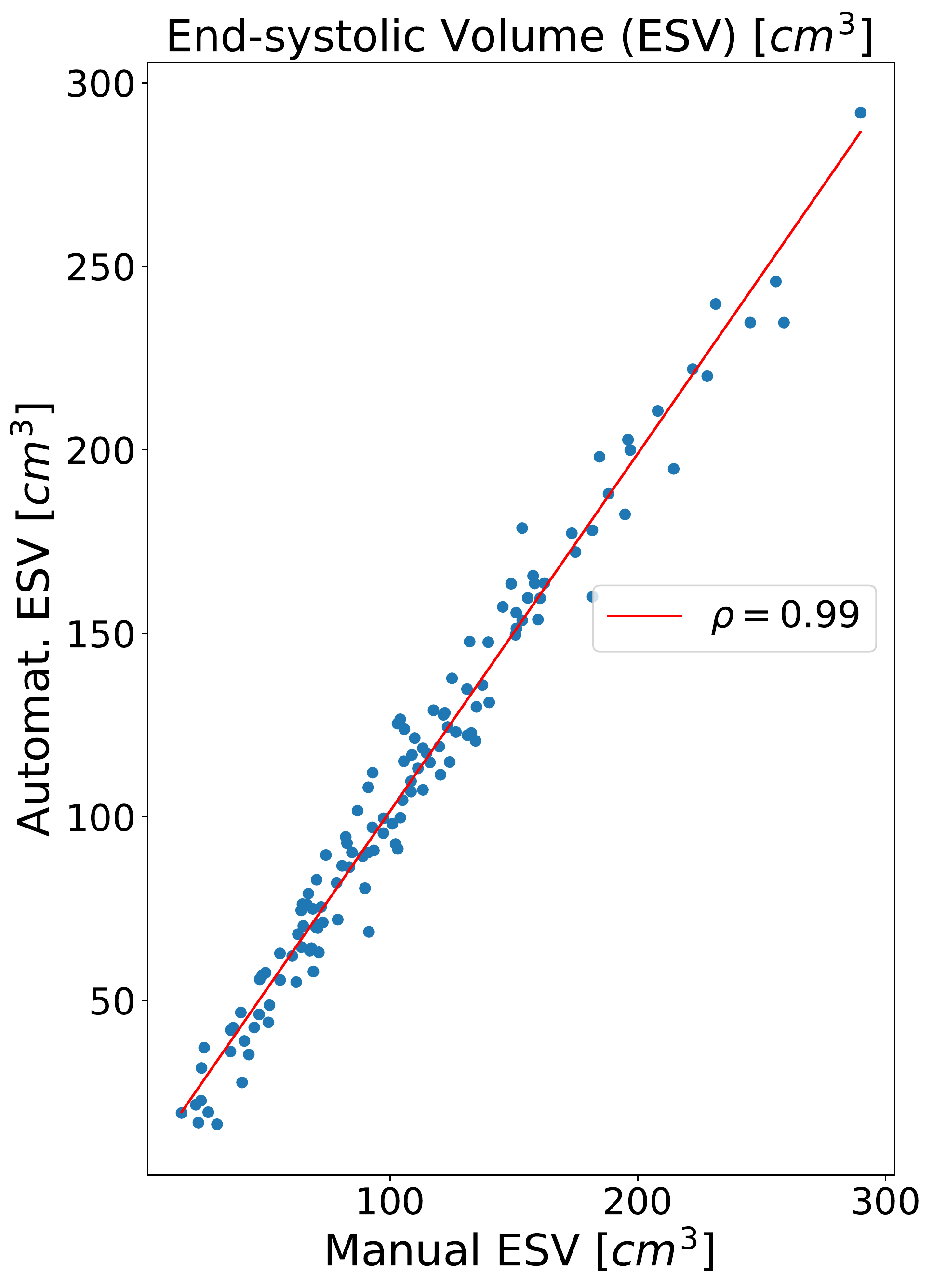}
\includegraphics[width=.315\textwidth]{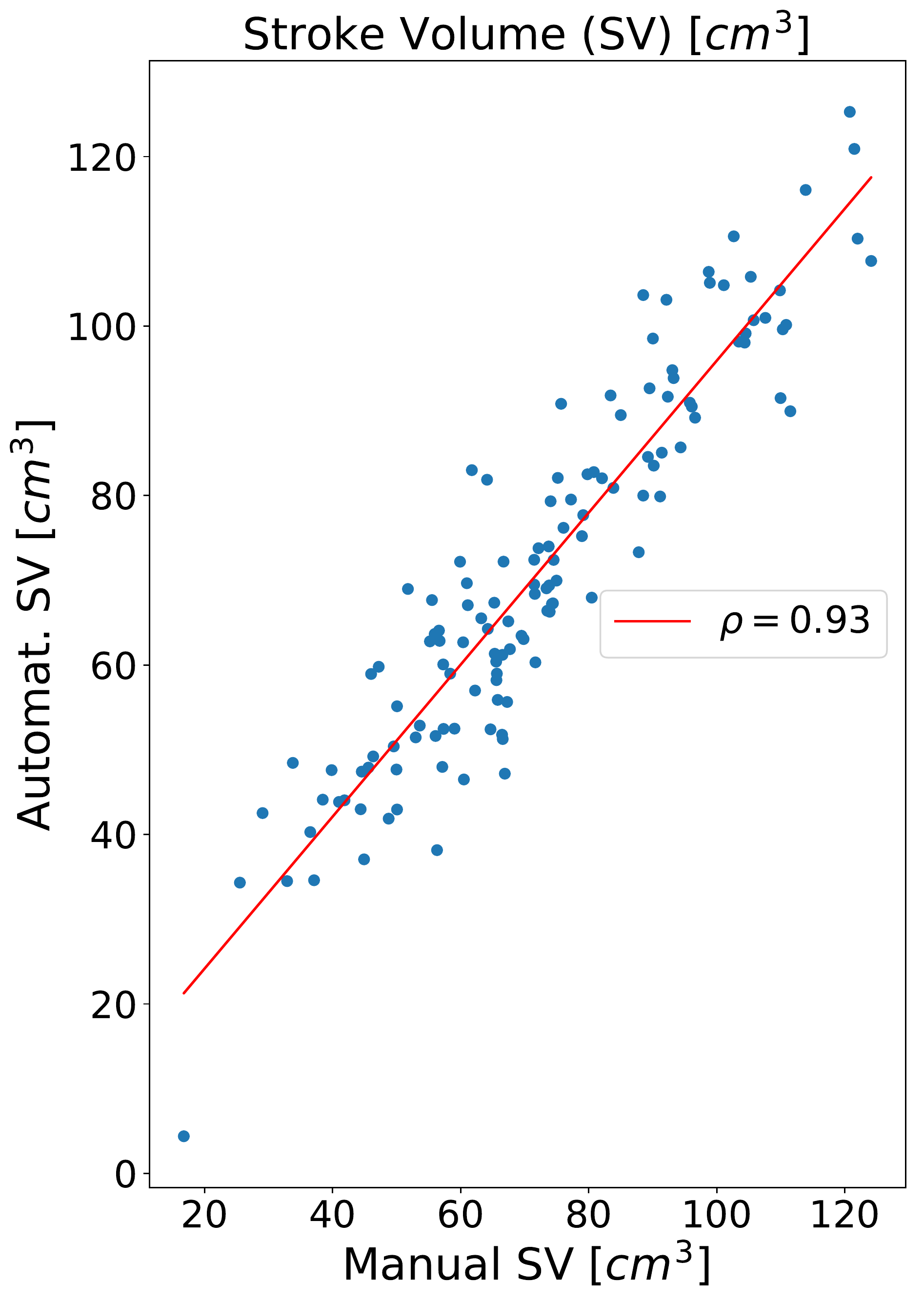}
\includegraphics[width=.32\textwidth]{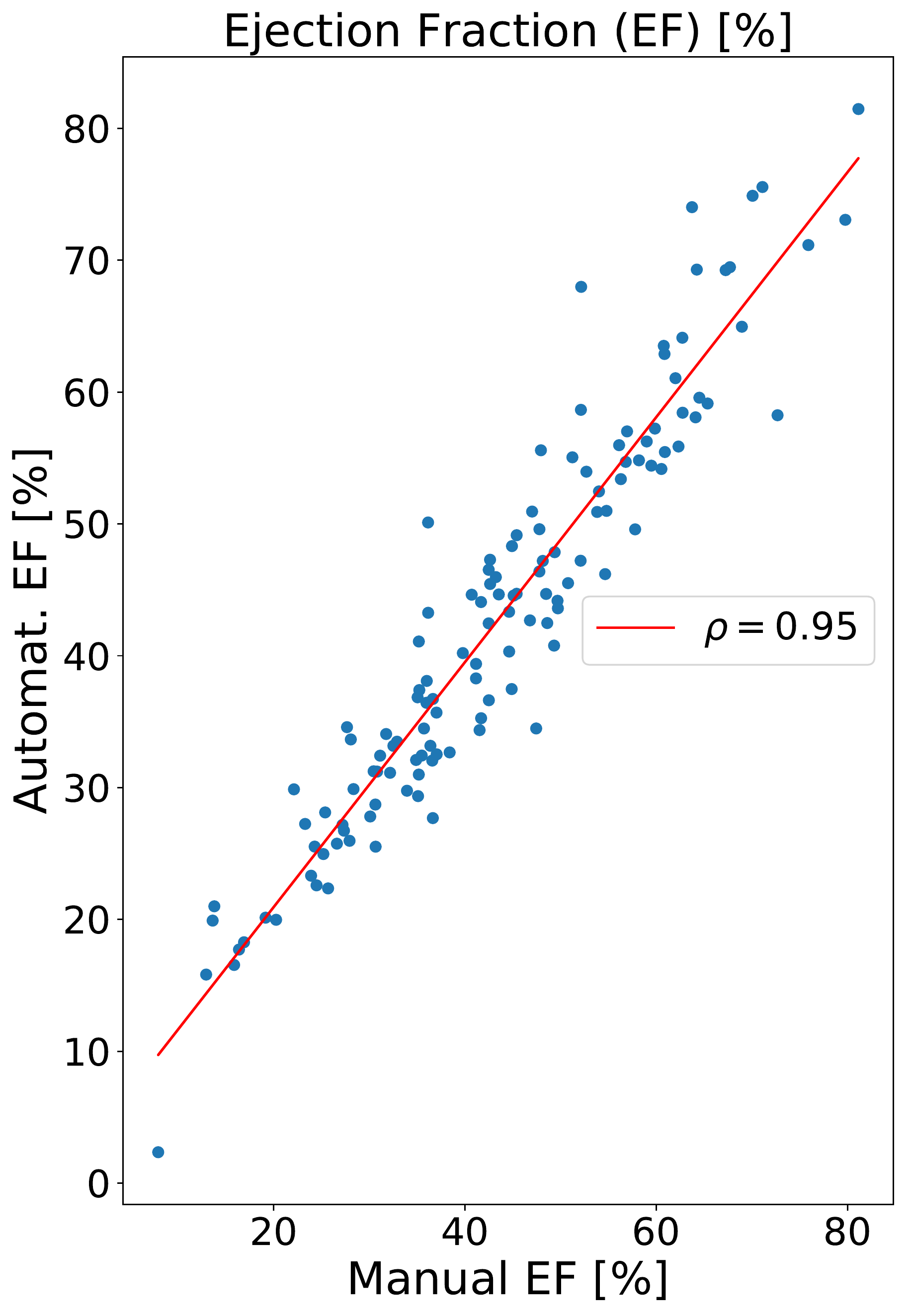}
\includegraphics[width=.328\textwidth]{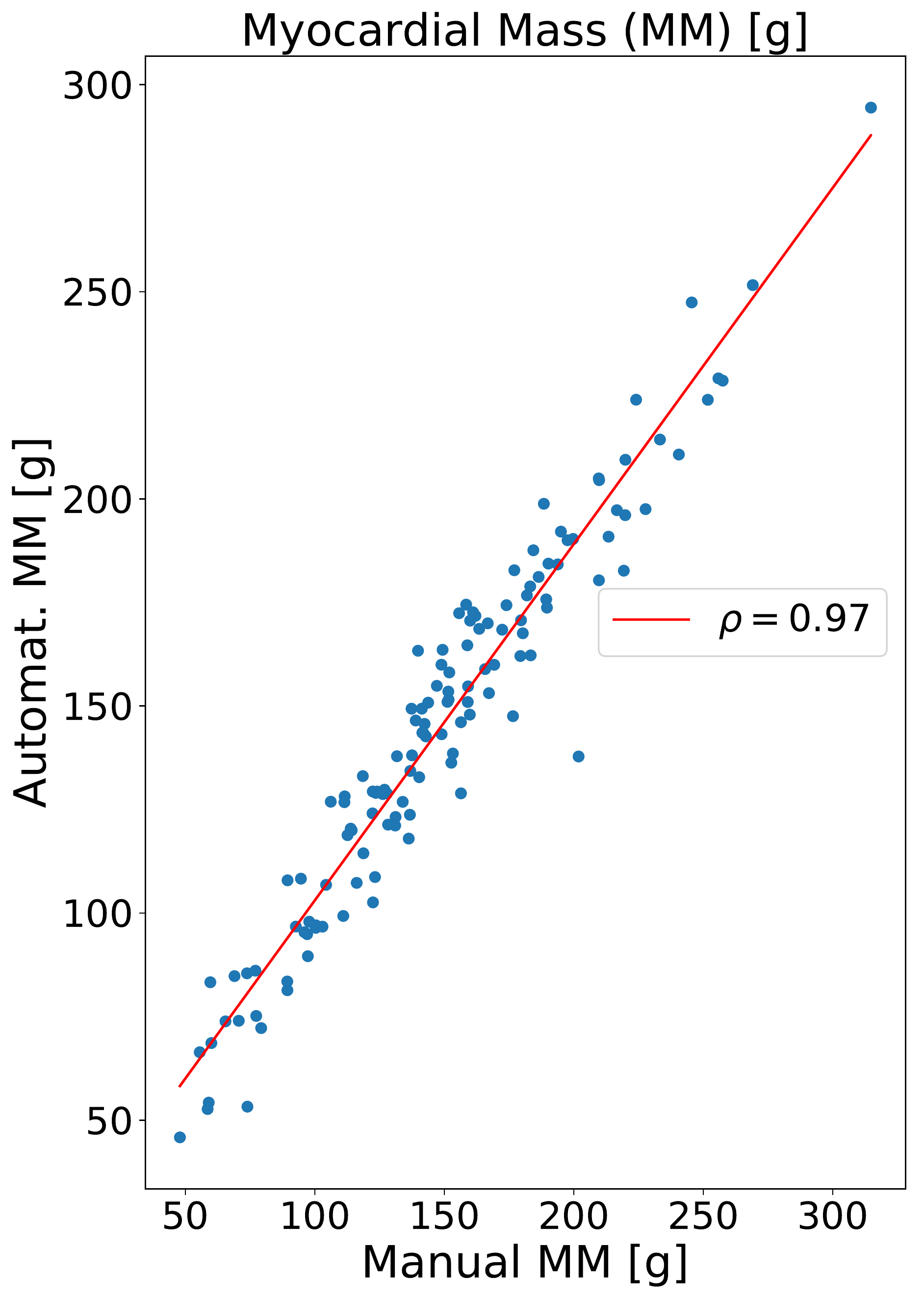}
\includegraphics[width=.31\textwidth]{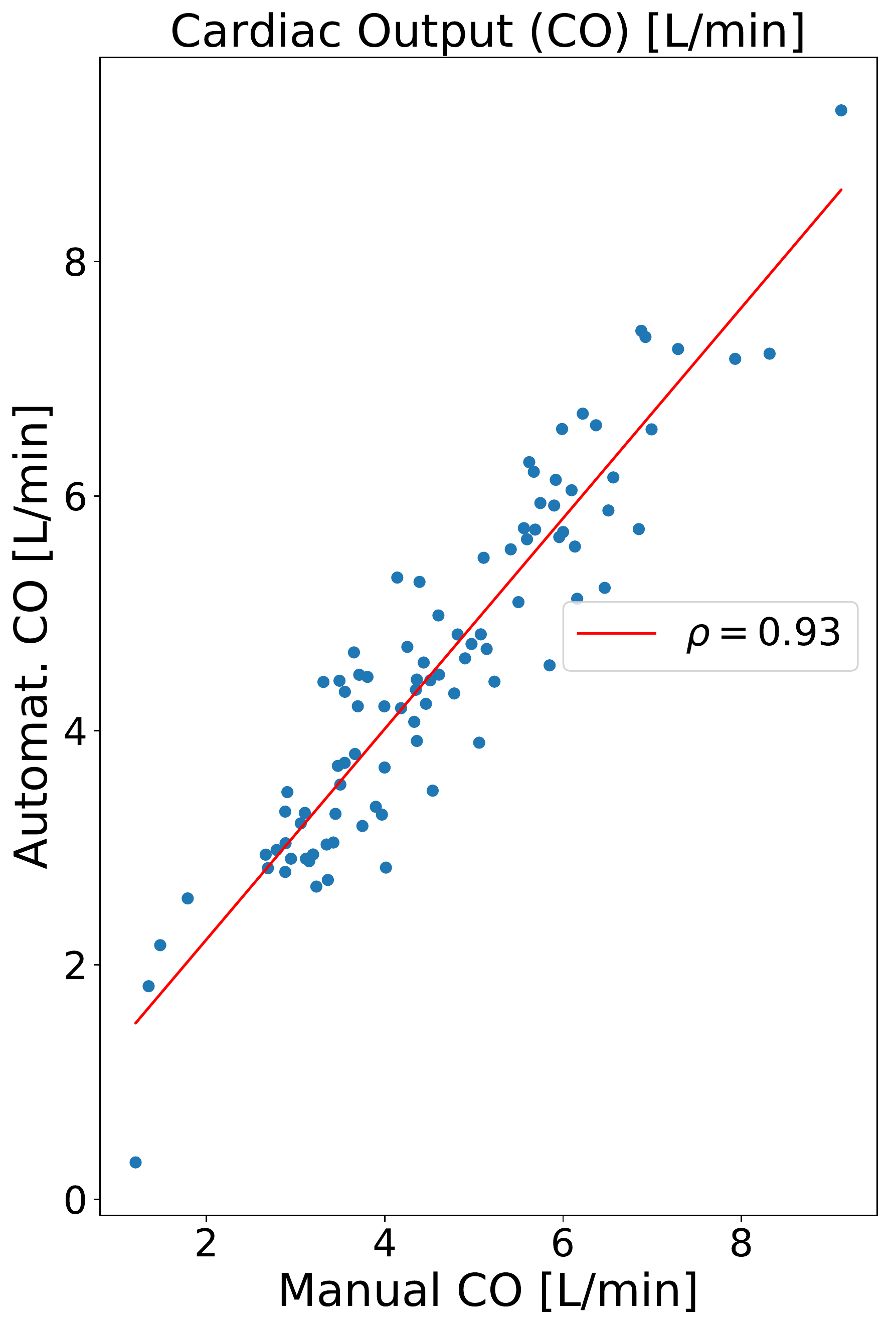}

\caption{Regression plots and the Pearson coefficient of the physiological measures studied. \label{fig_linear_regression}}
\end{figure*}

The Bland–Altman plots (Fig.~\ref{fig_blandaltman}) shows a minimal bias of $0.14\pm 0.51$~ml for EDV, a small bias of $1.37\pm 9.15$~ml for ESV, and $-0.71\pm 4.76$~\% for EF (95\% limits of agreement: -18.43 to 18.71 ml for EDV, -16.49 to 19.23 ml for ESV, and -10.02 to 8.59\% for EF). 
Plots of the linear regression of the measured parameters and those derived from the experts (Fig.~\ref{fig_linear_regression}) show a high correlation between the manual and automatic measures, specially  for the EDV and ESV. As it was described in the previous section, myocardial tissue and blood classification seems to be slightly more precise for end-diastole than for end-systole (see the standard deviation  and  the mean absolute erro in Table~\ref{table_phys_val}). Nevertheless, the overall effect of ESV inaccuracies in the EF calculation is negligible, and the proposed method is one of the most accuracy for measuring the EF. Also,  it is important to note that the EF and MM errors are comparable to the inter e intra-operator ranges for manual contouring reported in~\cite{Francescod}. The presence of outliers in the measures is mainly due to the segmentation of most basal and apical slices. The most basal slices are difficult to segment due to the myocardial tissue is not always complete in the slice and it changes the topology of the cavity. In the same way, the most apical slices are also difficult to segment because the blood pool is not always present in those slices which it changes the topology too.

\subsection*{Discussion}
\label{discussion}

\begin{figure*}[!t]
\centering
\includegraphics[width=.7\textwidth]{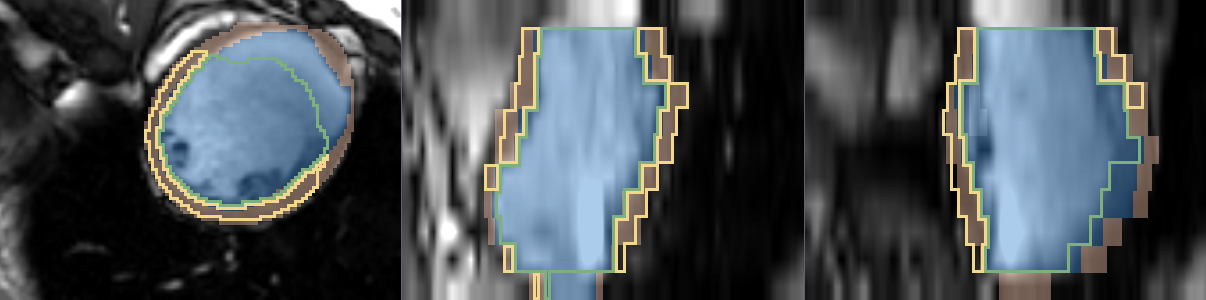}\\
\includegraphics[width=.7\textwidth]{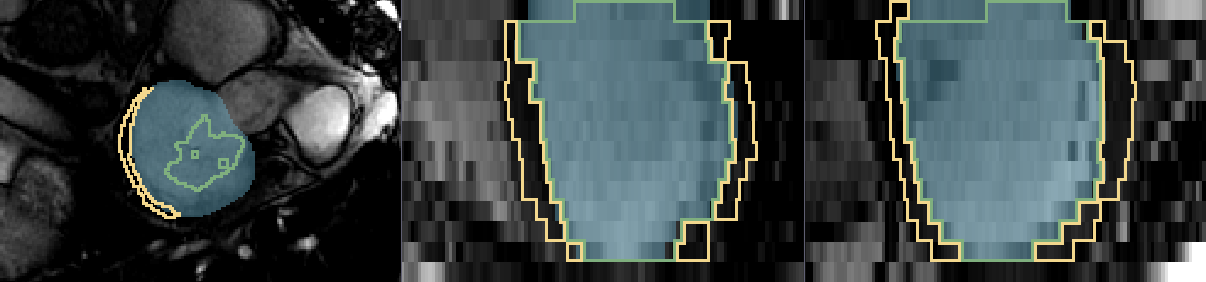}\\
\includegraphics[width=.7\textwidth]{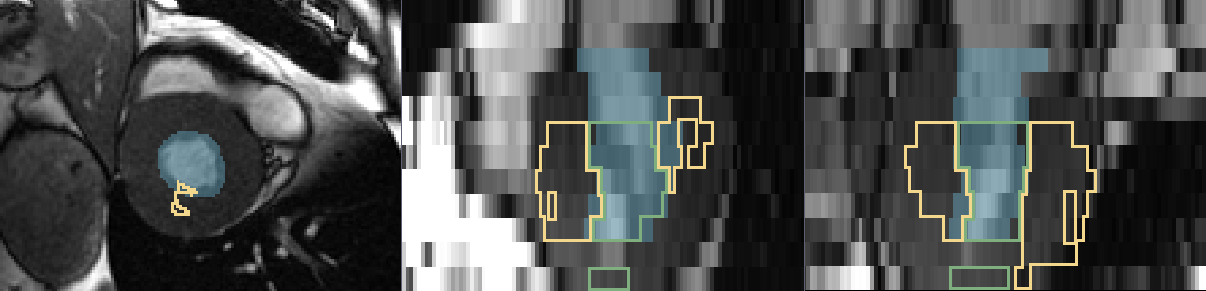}
\caption{Patients excluded from the physiological analysis. The automatic and manual segmentation is presented by patient (rows) in three orthogonal views (columns) at end-diastole (first row) and end-systole (middle and bottom rows). The blood pool is depicted in blue and green for manual and automated segmentation respectively. Also, the myocardial tissue is depicted in brown and yellow for manual and automated segmentation. The manual myocardial contour is only depicted when it is available (i.e. at end-diastole).\label{fig_pateints_fail}}
\end{figure*}

The ESV measures for three patients  were identified as  outliers with error values grater than 30~$cm^3$. These errors were found extremely higher with respect to others  patients. So, they were excluded from the previous analysis and they are discussed in what follows. An analysis on these patients reveals that the proposed approach for myocardial tissue classification fails (Fig.~\ref{fig_pateints_fail}). However,  is important to note that in one of these three patient, the method only fails  in a basal slice as it can be seen in Figure~\ref{fig_pateints_fail} middle row. The effect of this misclassification introduce an error of 49 and 46~$cm^3$ for EDV and ESV respectively. 

On the other hand, the proposed approach fails in different slices for the  other two patients as it is depicted in Figure~\ref{fig_pateints_fail}  first and bottom row. The  misclassification of the patient presented in the firs row results in a high error, but it is produced only at end-systole (30.5~$cm^3$). In this case, the patient presents a  severe dilated cardiomyopathy which made the proposed approach for tissue classification fails.
Similarly, the patient depicted in the bottom row in Figure~\ref{fig_pateints_fail} presents an hypertrophic cardiomyopathy. This hypertrophic myocardial tissue made the proposed approach reaches  an error  of 45.2~$cm^3$. Nevertheless, we believe that  the proposed approach will be able to overcome these misclassifications just by increasing the size of the dataset.

%
%

\section{Conclusions}
\label{conclusion}

In this paper, we have proposed an automatic LV function and mass quantification approach by  using deep learning networks. Unlike previous approaches, our method makes use of the Generalized Jaccard distance as objective loss function and residual learning strategies level to level to provide a suitable approach for myocardial segmentation and cardiac functional quantification. 
Quantitative and qualitative results show that the proposed approach presents a high potential for being used to estimate different structural and functional features for both prognosis and treatment of different pathologies. Thanks to data augmentation with free form deformations, it only need very few annotated images (280 cardiac MRI) and has a very reasonable training time of only  9 hours on a NVidia Tesla C2070 (6 GB) for reaching a  suitable accuracy of $\sim 0.9$ Dice's coefficient for both public datasets. Also, it is important to note that the methods achieves a strong correlation with  the most relevant functional measures (0.99 for EDV and ESV, 0.97 for myocardial mass, 0.95 for EF and 0.93 for SV and CO). And the error are comparable to the inter e intra-operator ranges for manual contouring. Additionally, this work leads to extensions for automatic detection and tracking of the right and left ventricle. Myocardial motion is useful in the evaluation of regional cardiac functions such as the strain and strain rate.

\section*{Acknowledgments}
This work was partially supported by Consejo Nacional de Investigaciones Cient\'ificas y T\'ecnicas (CONICET) and by grants M028-2016 SECTyP, Universidad Nacional de Cuyo, Argentina; and PICTO 2016-0023, Agencia Nacional de 
Promoci\'on Cient\'ifica y Tecnol\'ogica, and Universidad Nacional de Cuyo, Argentina.
German Mato acknowledges CONICET for the grant PIP 112 201301 00256.
\bibliographystyle{elsarticle-harv} 
\bibliography{IEEEabrv,abrv,refs}







\end{document}